\documentclass[10pt,twocolumn,letterpaper]{article}

\usepackage[pagenumbers]{cvpr} 

\usepackage[dvipsnames]{xcolor}

\usepackage{amsmath,amsfonts,bm}

\def\eqref#1{equation~\ref{#1}}

\def\1{\bm{1}}

\DeclareMathAlphabet{\mathsfit}{\encodingdefault}{\sfdefault}{m}{sl}
\SetMathAlphabet{\mathsfit}{bold}{\encodingdefault}{\sfdefault}{bx}{n}

\DeclareMathOperator*{\argmax}{arg\,max}

\usepackage[export]{adjustbox}

\definecolor{cvprblue}{rgb}{0.21,0.49,0.74}
\usepackage[pagebackref,breaklinks,colorlinks,citecolor=cvprblue]{hyperref}

\title{Comparing the Decision-Making Mechanisms by Transformers and CNNs 

via Explanation Methods}

\author{Mingqi Jiang, Saeed Khorram, Li Fuxin\\
Collaborative Robotics and Intelligent Systems (CoRIS) Institute\\
Oregon State University\\
{\tt\small   \{jiangmi, khorrams, lif\}@oregonstate.edu}
}

\begin{document}
\maketitle
\begin{abstract}
  In order to gain insights about the decision-making of different visual recognition backbones, we propose two methodologies, sub-explanation counting and cross-testing, that systematically applies deep explanation algorithms on a dataset-wide basis, and compares the statistics generated from the amount and nature of the explanations. These methodologies reveal the difference among networks in terms of two properties called compositionality and disjunctivism. Transformers and ConvNeXt are found to be more compositional, in the sense that they jointly consider multiple parts of the image in building their decisions, whereas traditional CNNs and distilled transformers are less compositional and more disjunctive, which means that they use multiple diverse but smaller set of parts to achieve a confident prediction. Through further experiments, we pinpointed the choice of normalization to be especially important in the compositionality of a model, in that batch normalization leads to less compositionality while group and layer normalization lead to more. Finally, we also analyze the features shared by different backbones and plot a landscape of different models based on their feature-use similarity.
  
\end{abstract}


\vspace{-0.06in}
\section{Introduction}
\vspace{-0.03in}
\label{sec:intro}
As attention-based Transformer networks 
show remarkable performance in image recognition tasks~\cite{dosovitskiy2020vit,liu2021Swin,graham2021levit,dai2021coatnet,xie2021segformer}, understanding and comparing Transformer and convolution networks (CNNs) at a deeper level become important. Prior work~\cite{URTIC,naseer2021intriguing} has  illustrated interesting differences between CNNs and transformers, but many questions have not been answered. Do transformers have  different inner working mechanisms? Why are some transformers seemingly more robust than CNNs? Recent work, such as ConvNeXt~\cite{convnext}, utilized design principles in transformer approaches to design a network based on depthwise convolutions and obtained excellent results. Does that indicate that the important contributing factor is not the attention itself but those design principles? If so, which specific design principles particularly affect the decision-making of networks?
Better answers to those questions would help us to gain more insights into those deep and complicated black-box networks. 

In this paper, we propose a novel methodology to examine these questions through \textit{deep explanation algorithms}. Explanation algorithms have improved significantly in recent years and can generate accurate explanations that can be verified through \textit{intervention experiments} on images~\cite{samek2016evaluating,2018RISE}. Recent search-based explanation algorithms can find a comprehensive set of \textit{minimally sufficient explanations} (MSEs)~\cite{shitole2021one}, defined as the minimal set of patches that, if shown to the network, lead to predictions that are almost as confident as predictions from the full image. The comprehensiveness of the set of MSEs produced by the search algorithm 
significantly surpasses 
traditional saliency maps that can only produce one explanation per image.

\begin{figure*}[t]
   \centering
    \includegraphics[width=0.85\linewidth]
    {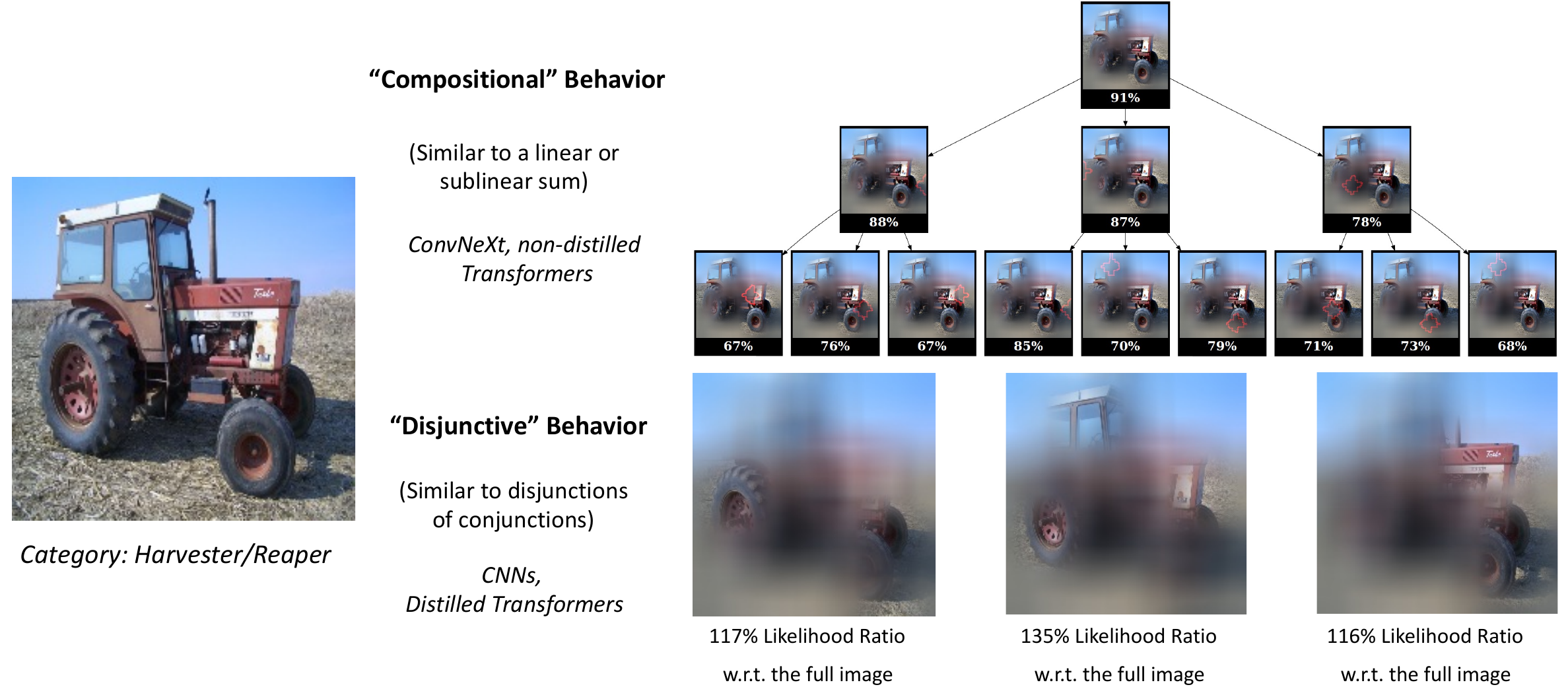}
   \vskip -0.15in
    \caption{\small Different behaviors exhibited by different classes of models. Likelihood ratio refers to the  ratio between the predicted class-conditional probability of the target category from the masked image and the full image. With the \textit{compositional} behavior, a confident classification is built up jointly from multiple parts, removing some parts may only slightly reduce the likelihood ratio (shown below each node in the tree in the top-right part of the figure). With the \textit{disjunctive} behavior, the network requires very few parts to obtain a highly confident prediction (sometimes more confident than the full image), but it can rely on any of multiple diverse combinations to obtain a confident prediction, similar to a logical OR among the different conjunctions (Best viewed in color)}
    \label{fig:teaser}
    \vskip -0.2in
\end{figure*}

While per-image explanation methods have greatly improved, they still do not provide a way to obtain a global understanding of the behavior of different network architectures. In this paper, we address this by extracting summary statistics from the explanations for each image and then combining them to obtain dataset-wide statistics.
With this approach, we hope to obtain insights that are no longer merely anecdotal, but statistically significant and verifiable.

We propose two approaches in this paper. The first is \textit{sub-explanation counting}, which investigates how networks perform on partial evidence by deleting patches from MSEs and examining the \textit{likelihood ratio} between the predicted conditional probability on those subsets of patches and the full image (Fig.~\ref{fig:teaser} Top-Right). The number of patch conjunctions that have high likelihood ratios indicates a type of behavior we call \textbf{compositional}, which means that the classification decision is built jointly on multiple local patches, and removing some of the patches merely lowers the confidence but may not change the classification decision.

We have observed significant differences across architectures regarding compositionality: 
ConvNeXt and transformer models without distillation are much more compositional, with \textbf{significantly more} subexplanations than regular CNN models. Further investigation showed that the most important factor in this difference, to our surprise, is not the choice between convolution and attention, but the \textbf{normalization} mechanisms used in the networks. Specifically, we found that the  \textbf{batch normalization} commonly adopted in CNNs leads to a significantly less compositional network, compared to the layer normalization commonly used in transformers. 
Receptive field size of the model also impacts compositionality to a lesser extent. 

The behavior of CNNs can be characterized as more \textit{disjunctive}, which means that the network can predict confidently from a smaller number of patches, although it can recognize any of several diverse patch combinations. Fig.~\ref{fig:teaser} Bottom-Right showed a few examples where a set of several revealed local patches lead to even more confident predictions than the full image, which reflects a distinctly different occlusion handling mechanism than compositional networks. We also found that commonly used \textbf{distillation} mechanisms that teach transformers with a CNN lead the transformers to become less compositional and more disjunctive, more similar to the CNNs. 

To address
the question whether different networks are using the same kind of visual features for classification, we developed a second methodology called \textit{cross-testing}.
In cross testing, we compute an explanation (image mask) for an image based on one network, and then submit the masked regions as input to the second network. 
This helps us to understand whether regions that contribute significantly to the first network are relevant to the second one. If two models rely on similar visual features, then they should score highly in cross-testing. On the other hand, if one model does not respond to the visual features that are deemed important to another model, 
this implies that
they are relying on different features. With this approach, we are able to plot the feature-use landscape of different convolutional networks and transformers, which demonstrates that different networks indeed use different features -- the cluster of CNNs, transformers and ConvNeXt are distinct from each other, although distillation can bring transformers closer to the CNNs.

In summary, our contributions are as follows:
\begin{itemize}
\item We propose two methodologies, subexplanation counting and cross-testing, which systematically apply model explanation approaches to examine the decision-making mechanisms of image recognition networks.
\item With sub-explanation counting, we revealed that the normalization layer significantly impacts model behavior -- batch normalization leads to disjunctive behavior (more combinations with fewer patches), while layer/group normalization leads to more compositional behavior (fewer combinations with more patches). Receptive field size also affects compositionality to a lesser extent.
\item With cross-testing, we are able to plot the feature-use landscape of different networks and show that CNNs, transformers and ConvNeXt do not use the same visual features for classification, whereas within each group the models are more similar to each other. 
\end{itemize}

\section{Related Work}
\noindent \textbf{Multiple Explanations.}
~\cite{ribeiro2018anchors} suggested multiple explanations might exist for the decisions made by the deep neural networks.~\cite{pmlr-v89-carter19a} proposed sufficient input subsets so that the observed values are sufficient to obtain output similar to the original input. They used instance-wise backward selection method to obtain such subsets.~\cite{shitole2021one} proposed \textit{Structured Attention Graphs} (SAG), which employs beam search to generate multiple 
sufficient
patch combinations. 

\noindent \textbf{Explanation Using Attribution (Heat) Maps.} 
Attribution maps (heatmaps) are some of the earliest and most widely-studied explanation tools for deep networks. They assign an \textit{attribution} score to each input feature that contributes to the desired output of the network. A majority of the early work, known as \textit{gradient-based} methods, generate attribution maps using the (modified) gradient of the output with respect to the input or intermediate features~\cite{MatDeconv, Gradcam17, JTguided2015, IntegratedGradient, smilkov2017smoothgrad, LRP15}. Later, sanity check procedures showed that most gradient-based explanation methods are independent of the model predictions and mainly work as edge detectors, greatly compromising their credibility \cite{adebayo2018sanity,2018Nie}. There are also concerns as to whether they are indeed interpretable by humans~\cite{zimmermann2021well}. 
\textit{Perturbation-based} approaches directly perturb the image regions (E.g., in \cite{ribeiro2016should} which works on superpixels). Most of such approaches optimize for a real-valued mask over the input features to find the regions that significantly decrease the output probability\cite{zintgraf2017visualizing,2018RISE, fong2019understanding}. However, 
optimization for a mask is highly non-convex and can be easily stuck in a bad local optimum. In addition, it is possible to generate \textit{adversarial} masks~\cite{hooker2019benchmark,adebayo2018sanity} that rely break
the input features to reduce output confidence. These are easy to locate but do not necessarily explain the decision-making of visual recognition models.
Recently, I-GOS \cite{IGOS} alleviated such issues by using the integrated-gradient as the descent direction rather than the gradient, which achieves faster convergence and locates better optima. They also proposed several tricks such as adding noise in the optimization process to avoid adversarial masks and retaining the masked image on the natural image manifold. iGOS++ \cite{khorram2021igos++} improved over \cite{IGOS} by additionally optimizing for minimal regions that improve the output confidence as well as enforcing a smoothness term inspired by bilateral filtering. This allowed faithful, non-adversarial and high-resolution masks to be discovered. 

\noindent \textbf{Understanding Transformers.}
Several works have explored the robustness of ViTs against CNNs under common perturbations \cite{URTIC, DBLP:conf/aaai/PaulC22, naseer2021intriguing}. \cite{naseer2021intriguing} observed that ViTs are significantly more robust to occlusions than ResNet50, with DeiT-S maintaining 70\% accuracy while ResNet50 drops to 0.1\% accuracy on ImageNet when 50\% of image regions are randomly removed. \cite{Mahmood_2021_ICCV} examined the adversarial robustness of ViTs. \cite{DBLP:conf/nips/RaghuUKZD21} studied the differences in the visual representations learned from ViTs and CNNs, particularly the utilization of global and local information across different layers. \cite{pmlr-v162-zhou22m,park2022how} further explored the role of self-attention in enhancing the robustness of vision transformers. ~\cite{park2022how} also revealed contrasting behaviors between attention and convolutional layers, where attention act as low-pass filters while convolutions function as high-pass filters. Different from previous work, our paper seeks to further analyze the underlying decision-making mechanisms transformers and CNNs use with explanation methods. 

\section{Methods}
\label{sec:method}
\subsection{Minimal Sufficient Explanations and Structural Explanations}
\cite{shitole2021one} showed that deep networks often have multiple ways to make classifications, and that a single explanation provided by heatmaps does not provide a complete understanding of the decision-making of the network. \cite{shitole2021one} proposed a more comprehensive way to find explanations by using beam search at low resolutions to systematically find different combinations of image regions that lead to high classification confidence for each image. 

Given a classifier $f$ that can predict $f_c(I) = \hat{p}(c|I)$ for an image $I$, we define the target class $\hat{c} = \argmax_c \hat{p}(c|I)$ as the class of the image predicted by the classifier. For simplicity, we also call $\hat{p}(\hat{c}|I)$ the \textit{classification confidence} of  $f$ on $I$. 
The goal is to examine whether the classification stays the same if the input is just a few patches of image $I$. For this goal, we can divide $I$ into non-overlapping patches $p_i$, usually at a low-resolution (e.g. $7\times 7$) to avoid adversarial perturbations to the image. Denote a union of those patches as a set $N$, we could predict the class-conditional probability $f_c(N) = p(c|I \cap N)$, denoting that only pixels in $N$ are retained while the rest are replaced with a baseline image of either $0$ or a blurred version of the original image~\cite{2018RISE}.

\begin{figure}
\begin{center}
\includegraphics[width=0.95\linewidth]{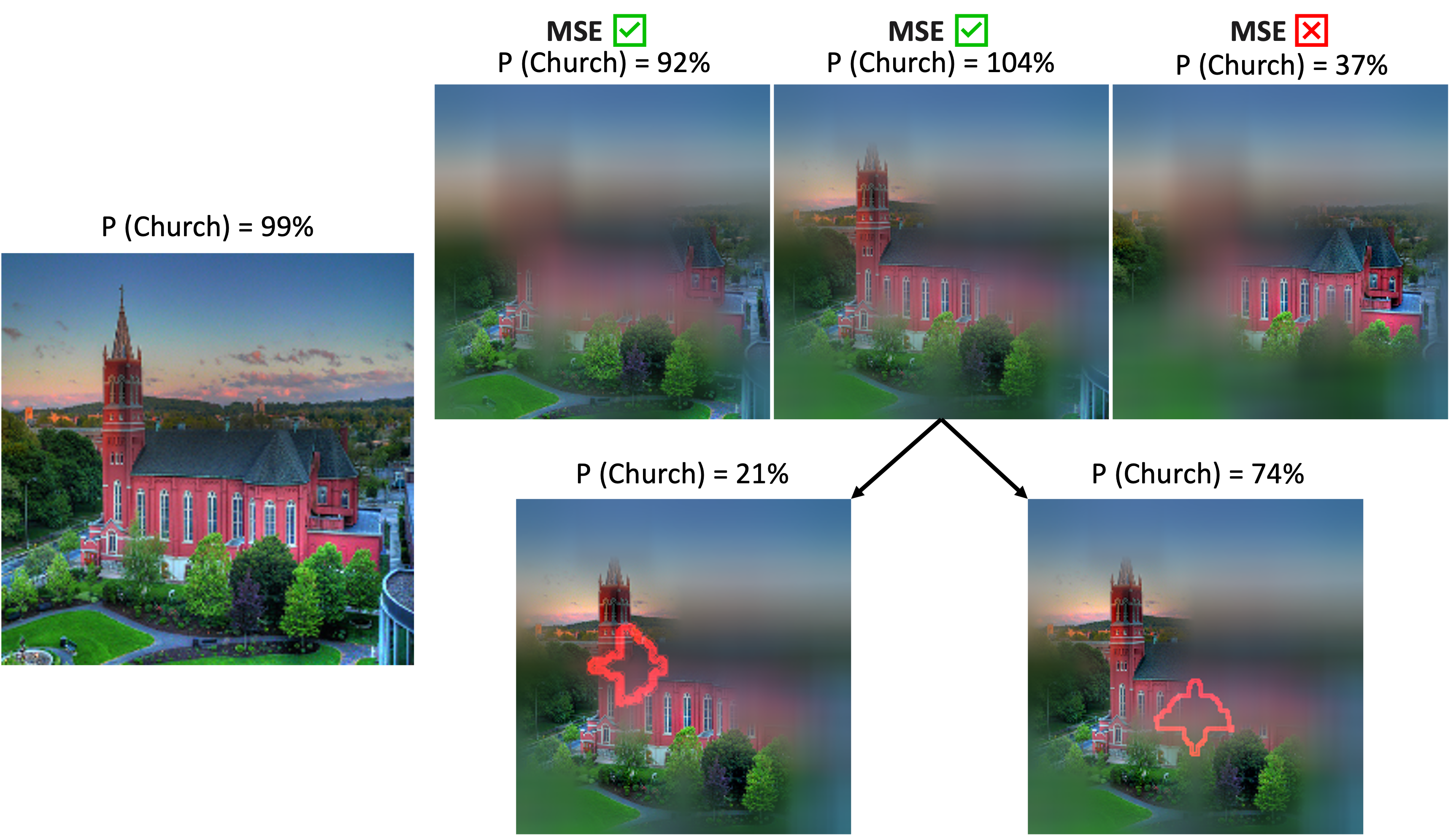}
\end{center}
\vskip -0.2in
\caption{\small Illustration of Minimal Sufficient Explanations (MSEs) and sub-explanations. MSEs are minimally masked images that the deep network would recognize as the same category as the full image, with its predicted class-conditional probability at least $90\%$ w.r.t. the one from the full image. Sub-explanations are defined as a subset of the patches of an MSE (Best Viewed in Color)}

 \vskip -0.2in 
\label{figure_mse_method}
\end{figure}

A Minimal Sufficient Explanation (MSE) is defined as a minimal set of patches that achieves a sufficiently high likelihood ratio w.r.t. the full image, i.e., $f_c(N) > P_{h}f_c(I)$, where $P_h = 0.9$ in their and our experiments. In layman terms, MSEs are the smallest region that, when shown to the deep network, can generate a prediction almost as confident as the whole image~(Fig.~\ref{figure_mse_method}). For simplicity, we will also call them \textbf{explanations} in the rest of the paper. MSEs are not unique and a \textit{beam search} can be used to efficiently find them. The search objective is to find all $N$s that achieve a likelihood ratio  higher than a threshold $P_h$, where no sub-regions in $n_j \subset N$ exceed that threshold,
{\small
\begin{align}
    f_c(N) \geq P_hf_c(I), \max_{n_j \subset N} f_c(n_j) < P_hf_c(I). \label{eq:MSE}
\end{align}
}

\subsection{Sub-Explanation Counting}

We propose to gain insights about different types of networks by counting \textit{sub-explanations}, defined as a subset of patches within an MSE: $N_s \subset N$ where $N$ is an MSE of an image $I$. By definition, $f_c(N_s) < P_h f_c(I)$, but there still can be two different types of behaviors: If the relationship among all patches is more similar to a logical conjunction (logical AND), then $f_c(N_s)$ could be quite low. However, deep networks are not necessarily logical, and there could be another type of behavior in that $f_c(N_s)$ still remains fairly high after occlusions of patches, which we define as a \textit{compositional} relationship (Fig.~\ref{fig:teaser} Top). Counting and comparing the number of sub-explanations across different models can help us understand which type of behavior each model is exhibiting.

Concretely, we construct a tree for each MSE by deleting one patch at a time from a parent node to generate child nodes. Every MSE for a given image is the root of a (sub)tree. In the meantime, we keep evaluating the confidence of current nodes (proper subsets of MSE) using the network $f$ and the image $I$ that is used to generate the MSE. When the nodes are with a likelihood ratio less than 50\% compared to the full image, we stop the expansion. Afterwards, we count the number of nodes that have classification confidence above several different thresholds. 

Note that being compositional is not the only way to be robust to occlusions. Instead, one could also have multiple MSEs in an OR-relationship to cover all possible  occlusions. In that way, the classifier still outputs  high classification confidence even with heavy occlusions as long as one of the MSEs corresponds to the occlusion pattern. In the experiments we contrast different models in this regard. 

\subsection{Metrics with Intervention Experiments} 
Next, we turn our attention to attribution maps (saliency maps), which is a very popular line of research in explanation but also controversial in that many of the algorithms have been shown to be unreliable~\cite{adebayo2018sanity}, mainly because earlier evaluation methodologies based only on localization were not necessarily correlated to the network classification. 
A better approach to evaluate the attribution map is via perturbing the input according to the map and evaluating the change in network prediction. \cite{samek2016evaluating} introduced \textit{MoRF} and \textit{LeRF} metrics in which the patches of image pixels are first ordered based on the attribution map values. Then, the most relevant features (MoRF) and least relevant features (LeRF) are gradually replaced by random noise sampled from a uniform distribution. Finally, the perturbed images are passed through the model and their classification confidences are obtained. Similarly, \cite{2018RISE} proposed the \textit{deletion} and \textit{insertion} metrics with the main difference being that during the perturbation, the substitute patches of pixels are sampled from a baseline image, e.g., a highly-blurred version of the image, rather than random noise. This way, sharp edges/boundaries are not introduced in the evaluation images, keeping them closer to the natural image distribution that the network  is trained on. 

\begin{figure}[tbh]
\includegraphics[width=0.8\linewidth, center]{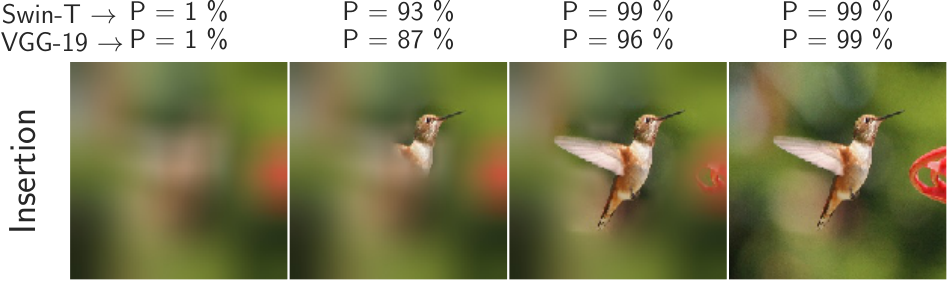}
\includegraphics[width=0.8\linewidth, center]{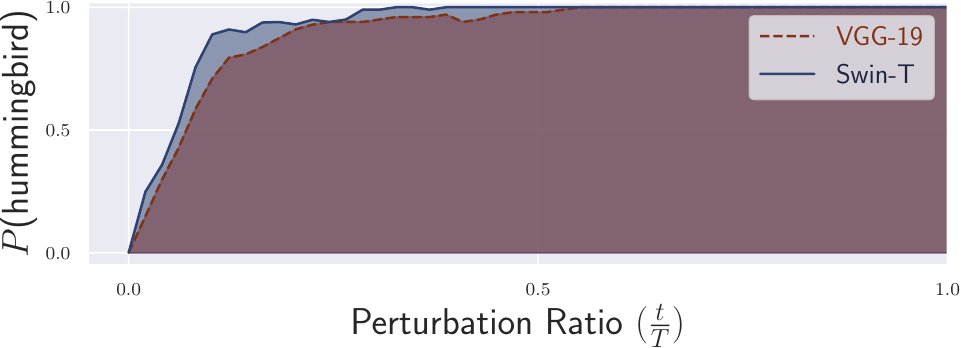} 
\vskip -0.1in 
\caption{\small Cross-testing the Insertion metric between VGG-19 and Swin-T for "hummingbird". (Top) Insertion images are obtained by successively revealing  pixels that are deemed salient by the heatmap; (Bottom) The Area Under the Curves (AUC) are used to compute the insertion metric for each classifier, when heatmaps are generated from only one of them (Best Viewed in Color)}
\vskip -0.1in 
\label{fig:ins_del}
\end{figure}

One can use the area under the curve (AUC) from the MoRF/deletion and LeRF/insertion curves as metrics reflecting the effectiveness of the explanation method in finding salient regions (Fig.~\ref{fig:ins_del}). In this paper we focus on the insertion metric, where a high insertion score indicates a sharp increase in the output confidence after the insertion of the most salient regions into the baseline image. Note that these evaluation schemes can be done automatically and do not require human-defined labels/bounding boxes \cite{zhang16excitationBP}, which makes large-scale quantitative evaluations easier. 
Formally, given an input image $I$, a baseline image $\tilde{I}$, classifier $f$, a target explanation class $c$, and an attribution map $M$ with elements in $[0,1]$, we can define the insertion metric as,
{\small
\begin{align} 
\label{eq:tr_metric}
    &\frac{1}{2T} \left<
    \sum_{t=0}^{T-1}f_{c}\left(\phi^{(t)}(\tilde{I}, I, M)\right) + f_{c}\left(\phi^{(t+1)}(\tilde{I}, I, M)\right)
    \right>_{p_{\text{data}}} \notag  \\
    &\phi^{(t)}(\tilde{I}, I, M) = I \odot M^{(t)} +  \tilde{I} \odot ({\bf 1} - M^{(t)})
\end{align}
}
where $T$ is the total number of perturbation steps, and $\phi^{(t)}$ generates the perturbed image after $t$ steps, i.e., $M^{(t)}$ only keeps the top $\frac{t}{T}$ of the pixels (Perturbation Ratio) in the attribution map and the rest of the pixels, if any, are set to zero. Of course, $\phi^{(0)}=I$ and $\phi^{(T)}=\tilde{I}$.

\subsection{Cross-Testing}
Inspired by the intervention-based metrics, we propose to evaluate the similarity between models by using one deep model to generate an attribution map, and successively mask the images based on the attribution map to assess the insertion/deletion metrics on the second  deep model. For fair comparison across different models which may have different average classification confidences, we normalize the scores based on the average top-1 classification confidence on the original image $t$ and  on fully blurred images $b$ for each model by $\bar{s} = {(s-b)}/{(t-b)}$~\cite{schulz2020restricting}. This method assesses the similarity of different models under occlusion. With a pairwise similarity matrix between two cross-tested models, we then utilize kernel-based dimensionality reduction approaches~\cite{kernelpca} to visualize them in a 2D space. This gives insights about whether the features used in the first network is salient to the second one.

\vspace{-0.05in}
\section{Experiments}\label{sec:exp}
\label{sec:experiment}

We compare ResNet50 \cite{ResNet}, ResNet50-C1, ResNet50-C2, ResNet50-D \cite{ResNet-strike},  VGG19 \cite{VGG19}, ConvNeXt-T \cite{convnext}, Swin-T \cite{liu2021Swin}, Nest-T \cite{Nest},  DeiT-S \cite{touvron2021training}, DeiT-S-distilled \cite{touvron2021training}, PiT-S~\cite{heo2021pit}, PiT-S-distilled~\cite{heo2021pit} and LeViT-256 \cite{graham2021levit} in our experiments. Of these, ResNet50 and VGG19 are older CNN models trained with less data augmentation. ResNet50-C1, ResNet50-C2 and ResNet50-D are ResNet50 variants trained with modern data augmentation strategies. ConvNeXt-T is a hybrid model based on large-kernel depthwise convolutions. 
Swin-T, Nest-T, DeiT-S, and PiT-S are transformers with different architectural structures.
DeiT-S-distilled, PiT-S-distilled and LeViT-256 were trained by distilling from a teacher CNN while DeiT-S and PiT-S were trained without distillation. The chosen models have similar sizes and similar accuracy on ImageNet (see Supplementary). To obtain standard deviations, we trained a few models with the same procedure but multiple random seeds, those results are provided in the Supplementary. 

In all experiments, we use the first 5,000 images from the ImageNet validation dataset \cite{imagenet_cvpr09} due to the slow speed of running all the experiments.

\begin{table}[tb]
\centering
  \resizebox{1.0\linewidth}{!}{%
  \begin{small}
  \begin{tabular}{cccccccc}
		\toprule
		\multicolumn{2}{c}{\bf Model} & \multicolumn{2}{c}{\bf MSEs} & \multicolumn{4}{c}{\bf Number of Subexplanations}\\
		\bf Type &\bf Name & \bf Count & \bf Size &   \bf $\geq$ 80\% & \bf $\geq$ 70\% & \bf $\geq$ 60\% & \bf $\geq$ 50\% \\
        \midrule
		  older CNNs  & VGG19  & 6.93 & 7.17 & 27.45 & 109.99 & 191.15 & 329.97\\
		     & ResNet50 & 6.76 & 7.28 & 53.68 & 108.55 & 180.44 & 296.92\\
      \midrule
		   & ResNet50-C1 & 9.52 & 6.37 & 194.16 & 320.53 & 430.73 & 591.69 \\
		 newer CNNs  & ResNet50-C2 & 11.01 & 5.94 & 88.82 & 202.27 & 369.35 & 568.91 \\
		   & ResNet50-D & 9.88 & 6.02 & 146.78 & 216.31 & 272.459 & 332.22\\
    \midrule
		  ConvNeXt  & ConvNeXt-T  & 10.28 & 6.14 & \bf 980.16 & \bf 2001.67 & \bf 3610.37 & 5360.43\\
		\midrule
            & Swin-T & 8.90 & 8.01 & 221.58 & 882.72 & 2933.03 & \bf 7268.20\\
		Transformers      & Nest-T & 7.18 & \bf 8.77 & 432.37 & 1093.08 & 2725.06 & 6006.22 \\
               & DeiT-S & 8.95 & 7.72 & 72.09 & 333.84 & 1097.58 & 2408.30 \\
      & PiT-S & 7.89 & 7.49 & 131.32 & 607.97 & 1803.04 & 3862.10 \\
      \midrule
	Transformers	       & DeiT-S-dis & 10.22 & 5.77 & 57.52 & 114.21 & 227.41 & 467.72 \\
              with  & PiT-S-dis & 10.06 & 5.86 & 48.54 & 91.45 & 182.67 & 334.45 \\
		     Distillation  & LeViT-256 & \bf 12.59 & 5.50 & 54.96 & 103.24 & 177.33 & 253.66 \\
		\bottomrule
  \end{tabular}
  \end{small}
  }
  \vskip -0.1in
  \caption{\small Results of beam search to locate MSEs. The numbers on the top right are thresholds on the likelihood ratio between a subexplanation and the full image}
  \label{tab:example4}
  \vskip -0.2in
\end{table}

\subsection{Explanations and Sub-Explanations}
\label{sec:experiment_MSEs}
We follow \cite{shitole2021one} to perform a beam search with width $5$ on different patch combinations,  with the image divided into a $7 \times 7$ grid with $49$ patches. 
The baseline image was set to a blurred version of the original image (see Supplementary for  results with a zero-image baseline).
In Table~\ref{tab:example4}, we count the number of MSEs and subexplanations among different networks. 

\noindent \textbf{Disjunctivism and Compositionality.} Table~\ref{tab:example4} shows distinct differences among the different models. Most CNNs, ConvNeXts and distilled transformers have  \textbf{higher} MSE counts and  \textbf{smaller} MSE sizes. In contrast, Swin Transformers and other undistilled transformers have smaller MSE counts and larger MSE sizes. The differences are statistically significant (tests shown in the Supplementary). 

\begin{figure*}[htb]
\centering
\includegraphics[width=0.86\linewidth]{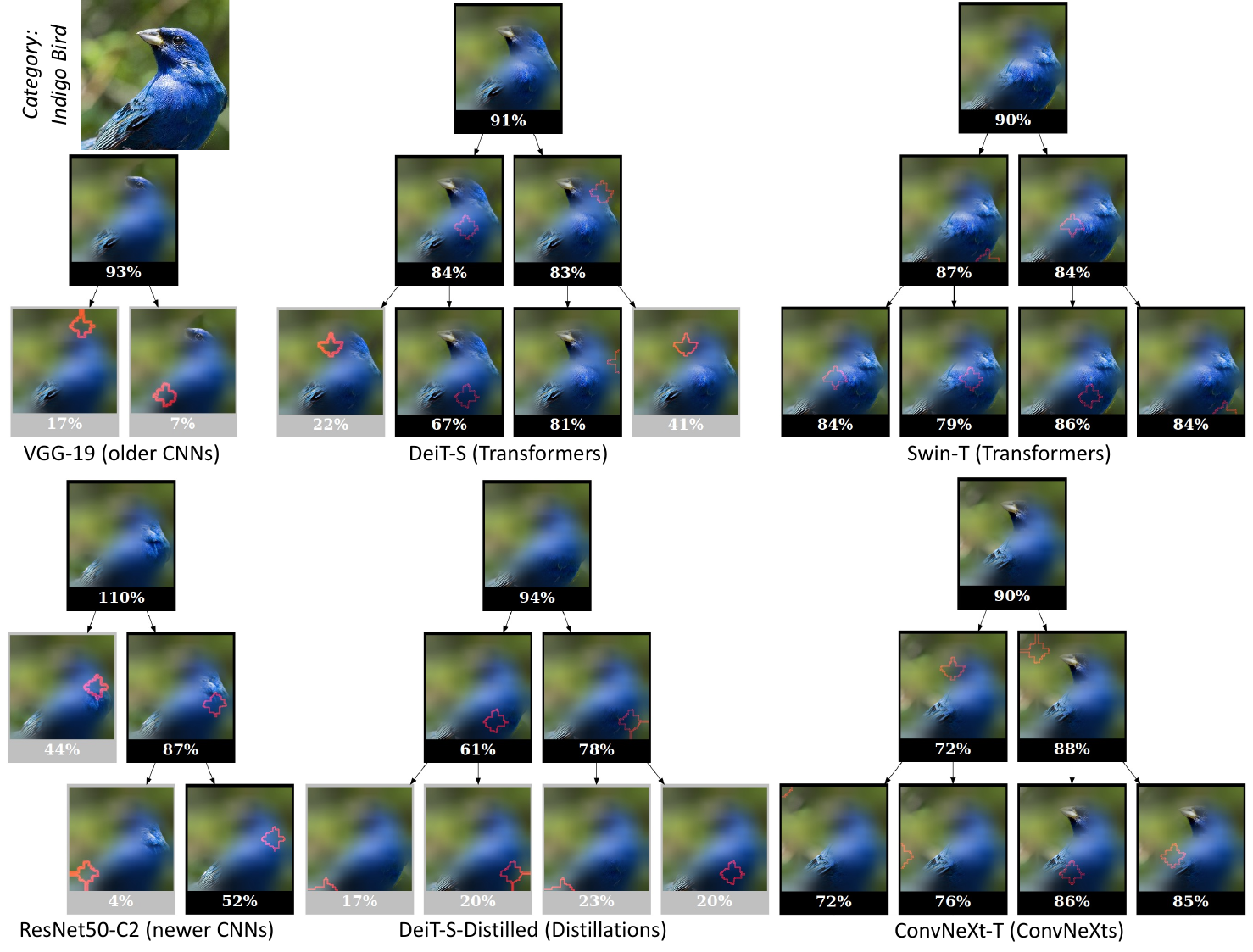}
\vskip -0.15in 
\caption{\small MSEs and some sub-explanations of different models on an image of the Indigo Bird class. Due to the space limit we only subsampled a few subexplanations. The removed patch from the parent node is indicated with a red outline. 
(Best viewed in Color)
}
\label{figure_tree_examples}
\vskip -0.2in 
\end{figure*}

Recalling the definition of MSEs, a higher count and smaller size means that the network  needs the conjunction of \textbf{fewer} patches to form a confident classification. However, the network can be robust to occlusions or missing visual features, since it can use a \textbf{different} conjunction if a certain important feature cannot be seen. This is what we define as disjunctivism. 

In contrast, for transformer models which exhibit larger MSE sizes, we note that the number of subexplanations is also  significantly higher. This suggests the \textbf{compositional} mechanism for handling occlusions: in each conjunction of patches, removing some of the patches only slightly lowers the classification confidence, whereas in CNNs and distilled transformers removing some of the patches in an MSE greatly lowers the classification confidence (leading to fewer subexplanations). It is clear that disjunctivism and compositionality are different mechanisms that can both deal with occlusion and missing features.

A separate result is the effect of data augmentation: newer CNNs with better data augmentation has significantly more MSEs than older ones, showing higher robustness.

A disclaimer is that these are overall trends that can only be observed by systematically evaluating the explanations on a large dataset. We can find any network using any of these inference strategies for a specific image, and in many images all networks use a similar set of features, leading to smaller differences overall among them in Table~\ref{tab:example4} (see Fig.~\ref{figure_tree_examples} and Supplementary for visual examples). This shows the importance of the statistical approach we are taking w.r.t. explanation methods, as it uncovers the trend from the noisy signals of individual images.

\noindent \textbf{What drives the high number of sub-explanations in ConvNeXt and Swin Transformers?}
One specifically interesting aspect is the \textbf{significantly higher} number of subexplanations in ConvNeXt and transformers without distillation. For example, the subexplanations with $\geq 50\%$ confidence ratio are usually in the thousands in those networks, versus hundreds in the CNNs and transformers with distillations. ConvNeXt especially, shows up as an outlier in our analysis as it looks \textbf{both} having more MSE counts and smaller sizes similar to CNNs as well as being compositional with more subexplanations. Hence, we set out to examine which design aspect specifically drove the high number of sub-explanations. 

We attempted to strip out the design elements of ConvNexT one by one, and mirrored the experiment with Swin Transformers as well.
Specifically, we trained ConvNeXt-T using a 3×3 kernel size for all the ConvNeXt blocks, and Swin-T using a 4x4 window size in the first two stages, and name the resulting models \textit{ConvNeXt-T-3} and \textit{Swin-T-4}. Noting that the results did not fully explain the differences in subexplanations, we replaced the original layer normalization (LN) with batch normalization (BN) and group normalization (GN) and further trained models with different normalizations and smaller receptive fields: 
\textit{ConvNeXt-T-3-BN}, \textit{ConvNeXt-T-3-GN}, \textit{Swin-T-4-BN}, and \textit{Swin-T-4-GN}. 
These changes did not reduce performance on ImageNet. More information is provided in the Supplementary.

\begin{table}
\centering
  \resizebox{1.0\linewidth}{!}{%
  \begin{small}
  \begin{tabular}{cccccccc}
		\toprule
		\multicolumn{2}{c}{\bf Model} & \multicolumn{2}{c}{\bf MSEs} & \multicolumn{4}{c}{\bf Number of Subexplanations}\\
		\bf Type &\bf Name & \bf Count & \bf Size &   \bf $\geq$ 80\% & \bf $\geq$ 70\% & \bf $\geq$ 60\% & \bf $\geq$ 50\% \\
        \midrule
		    & ConvNeXt-T  & \bf 10.28 & 6.14 & \bf 980.16 & \bf 2001.67 & 3610.37 & 5360.43\\
        ConvNeXts    & ConvNeXt-T-3 & 9.31 & 6.56 & 526.40 & 1136.17 & 2012.84 & 3089.83 \\
            & ConvNeXt-T-3-GN & 6.60 & \bf  7.96 & 471.87 & 1468.28 & \bf 3742.13 & \bf 7476.92\\
            & ConvNeXt-T-3-BN & 9.31 & 6.50 & 64.39 & 157.46 & 326.03 & 672.92 \\
		\midrule
            & Swin-T & 8.90 & 8.01 & \bf 221.58 & \bf 882.72 & \bf 2933.03 & \bf 7268.20\\
        Transformers  & Swin-T-4 & 8.11 & 7.46 & 139.29 & 588.75 & 1885.98 & 4276.08 \\
         & Swin-T-4-GN & 6.57 & \bf 9.42 & 207.58 & 821.59 & 2641.81 & 7039.34 \\
            & Swin-T-4-BN & \bf 9.40 & 7.18 & 42.92 & 127.05 & 387.41 & 943.29 \\
		\bottomrule
  \end{tabular}
  \end{small}
  }
  \vskip -0.1in
  \caption{\small Results of beam search to locate MSEs on ConvNeXt and Swin variants. The numbers on the top right are thresholds on the likelihood ratio between a subexplanation and the full image}
  \label{tab:sub_ablation}
  \vskip -0.22in
\end{table}

The results in Table~\ref{tab:sub_ablation} are quite \textbf{surprising} for us, as we did not expect batch normalization to play such a significant role: reducing the size of the receptive field reduced the subexplanations by about $40\%$, but changing layer normalization to batch normalization \textbf{very significantly} reduced the number of subexplanations by about \textbf{80\%}, driving ConvNeXt and Swin Transformer back to  levels similar with CNNs. 
This shows that although receptive field size and normalization both played a significant role in compositionality, the choice of normalization is a much stronger factor. GN exhibited compositional behaviors similar to LN, and both are distinctly different from BN.

\begin{figure}
\centering
\includegraphics[width=0.7\linewidth]{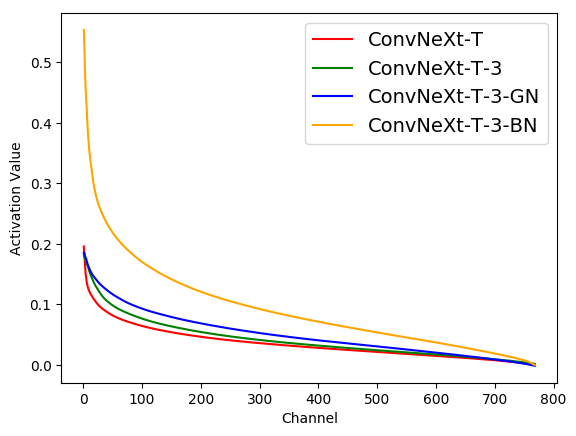}
\vskip -0.15in 
\caption{\small Sorted average values of the maximal activation in each image for each channel in the last block for ConvNeXt-T variants}
\label{activation_convnext}
\vskip -0.2in 
\end{figure}

Put in other words, using \textbf{BN strongly} leads the network to be  \textbf{less compositional}, in the sense that missing features in a conjunction drops the prediction confidence more quickly. This makes the relationship among features more like logical AND/ORs, rather than a linear sum. We attempt to explain this by examining the normalization dimension of these approaches. Batch normalization only normalizes within a single channel and \textbf{does not} normalize across different network channels, whereas GN and LN normalize across different channels. This could lead to an effect that a few large activations dominate the prediction when a network is using BN. Fig.~\ref{activation_convnext} shows the activation map values in different networks and indeed the trend is clear that the top feature channels when using BN are much more dominant than with GN and LN. This effect is a preference during the network training process -- it does not mean that BN and LN lead to fundamentally different network architectures. Table~\ref{tab:example4} showed that distillation from a CNN can 
reduce the number of subexplanations of  an LN-normalized transformer and thus reduce its compositionality.

Having discovered the effect, the open question is whether it is good or bad? We do not have enough concrete evidence to support an outright answer, but some intuitive argument for compositionality can be made -- over-reliance on the existence of a few local features might reduce robustness to adversarial examples. Although disjunctivism compensates by introducing more conjunctions, the increase is only about 2 MSEs per image on average.
Calibration under uncertainty might also be easier with compositionality, because it is easier to generate ``semi-confident" predictions. ConvNeXt and Swin Transformers usually dominate distilled ViTs on downstream visual tasks such as detection and segmentation, where one can argue it might be important to look at more features, not only the ones that are most discriminative. Potentially, one could also take a harder look at combinations of batch and group normalizations such as \cite{zhou2020batch}, which indeed showed better adversarial accuracy and domain adaptation capabilities but was hardly ever used in recent  architectures. A potential argument for disjunctivism might be to support more consistent predictive confidence under occlusions, which could be useful if a firm decision needs to be made regardless of occlusions.

\subsection{Cross-Testing with Attribution Maps}
\label{sec:cross_testing_with_Heatmap}

\begin{figure}[htb]
\vskip -0.12in
\centering
\includegraphics[width=0.9\linewidth]{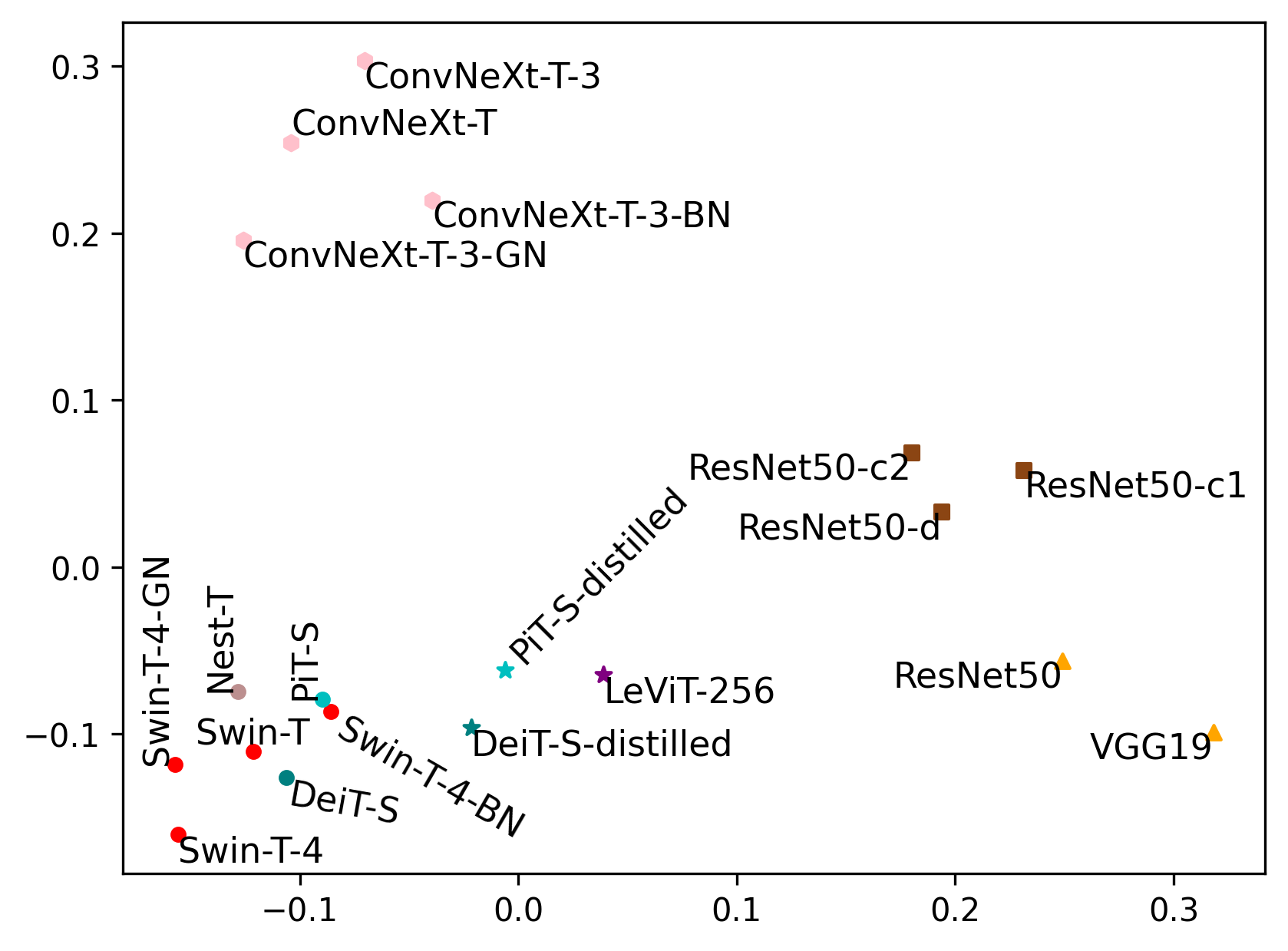}
\vskip -0.15in
\caption{\small Kernel PCA projections of different models using the insertion metrics. 
We can see that models that are the same type are more similar to each other in this plot, and that distillation brings transformers closer to CNNs}
\vskip -0.1in
\label{figure_pac}
\end{figure}

The other question we seek to answer is whether different types of networks are using similar features to classify via cross-testing their attribution maps.
We use a state-of-the-art attribution map method iGOS++ \cite{khorram2021igos++} to generate heatmaps for each image at $28 \times 28$ resolution. This resolution is chosen because it is the highest resolution for which iGOS++ has consistently good performance across different networks. We then calculate the insertion and deletion scores based on the obtained heatmap values (full results in Supplementary). In order to better visualize these similarities, we applied Kernel PCA to project them to $2$ dimensions \cite{kernelpca}, based on the similarities of the insertion scores. Figure \ref{figure_pac} shows the projection results. It can be found that the same type of models use similar features for their predictions. We can roughly delineate clusters  of older CNNs (VGG19, ResNet50), newer CNNs (ResNet50-C1, ResNet50-C2, ResNet50-D), ConvNeXt variants (ConvNeXt-T, ConvNeXt-T-3, ConvNeXt-T-3-BN and ConvNeXt-T-3-GN), non-distilled transformers (Swin-T, Swin-T-4, Swin-T-4-BN, Swin-T-4-GN, Nest-T, Deit-S, PiT-S) and distilled transformers (PiT-S-distilled, Deit-S-distilled and LeViT-256).
Results using iGOS++ with setting perturbed pixels to zero 
and another attribution map approach, Score-CAM \cite{wang2020score}, are shown in the supplementary which show similar trends.

\begin{figure*}[htb]
        \center
        \begin{minipage}{0.9\linewidth}
        \center
        \setlength\tabcolsep{0.5pt}
        \begin{tabular}{c|c|c|c|c|c|c|c|c|c|c|c|c|c|c|c|c|c|c|c|c|c|c|c}
        
                \multicolumn{6}{c}{\small Sea Snake} & \multicolumn{6}{c}{\small Bakery} & \multicolumn{6}{c}{\small Spoonbill} & \multicolumn{6}{c}{\small Dial Phone} \\
        
                \multicolumn{6}{c}{\includegraphics[width=0.24\linewidth,height=2.5cm]{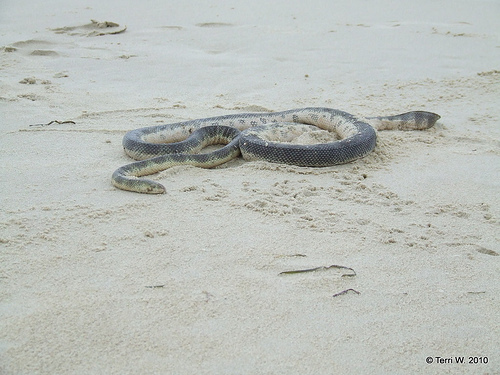}} &    \multicolumn{6}{c}{\includegraphics[width=0.24\linewidth,height=2.5cm]{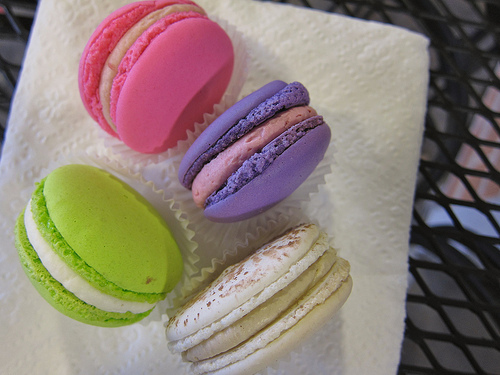}} &
                \multicolumn{6}{c}{\includegraphics[width=0.24\linewidth,height=2.5cm]{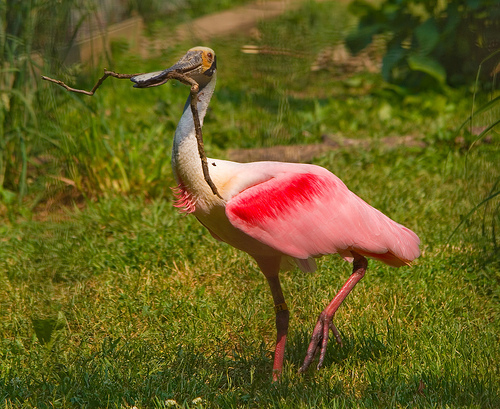}} &
                \multicolumn{6}{c}{\includegraphics[width=0.24\linewidth,height=2.5cm]{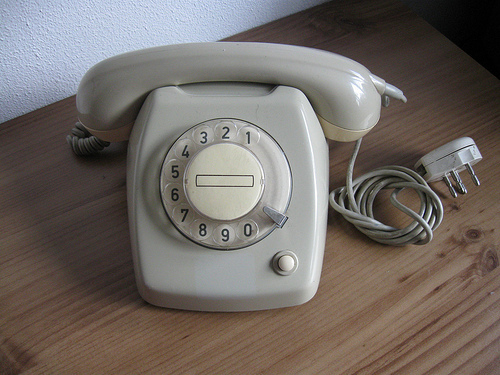}} \\

                \multicolumn{6}{c}{\includegraphics[width=0.24\linewidth,height=2.5cm]{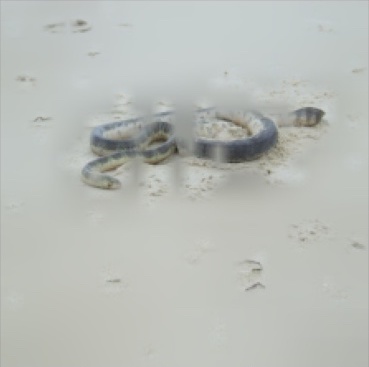}} &
                \multicolumn{6}{c}{\includegraphics[width=0.24\linewidth,height=2.5cm]{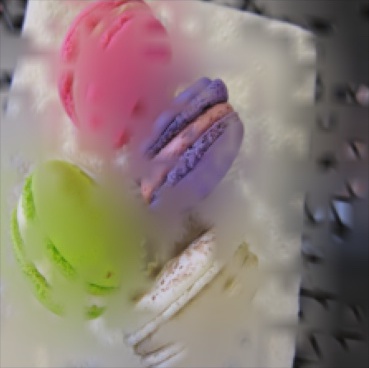}} &
                \multicolumn{6}{c}{\includegraphics[width=0.24\linewidth,height=2.5cm]{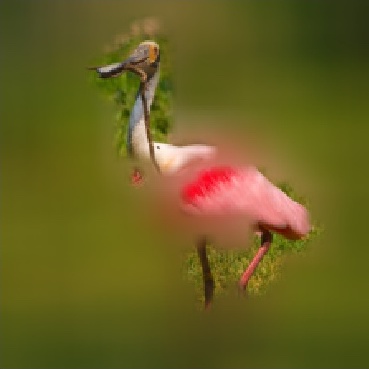}} &
                \multicolumn{6}{c}{\includegraphics[width=0.24\linewidth,height=2.5cm]{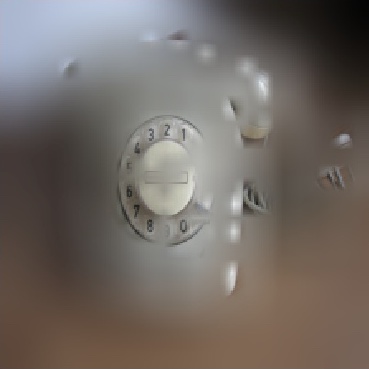}}\\
                
        \end{tabular}
        \end{minipage}

        \vskip -0.05in
        \begin{minipage}{0.88\linewidth}
        \center
        \scriptsize
        \resizebox{1\linewidth}{!}{%
        \begin{small}
        \begin{tabular}{@{ }c@{ }|@{ }c@{ }|@{ }c@{ }||@{ }c@{ }|@{ }c@{ }|@{ }c@{ }||@{ }c@{ }|@{ }c@{ }|@{ }c@{ }||@{ }c@{ }|@{ }c@{ }|@{ }c@{ }}
                \multicolumn{12}{c}{Prediction Confidence on the Partially Occluded Image}\\

                \toprule
             VGG19 & ResNet50-c2 & ConvNeXt-T &
                VGG19 & ResNet50-c2 & ConvNeXt-T &
                VGG19 & ResNet50-c2 & ConvNeXt-T &
                VGG19 & ResNet50-c2 & ConvNeXt-T \\
                \midrule
                0.0494 & 0.2988 & 0.5609 & 
                0.0775 & 0.4752 & 0.1447 &
                \bf 0.9825 & 0.8261 & 0.9027 &
                0.0043 & 0.0607 & 0.0595 \\
                
                \midrule

                DeiT-S & DeiT-S-dis & Swin-T &
                DeiT-S & DeiT-S-dis & Swin-T &
                DeiT-S & DeiT-S-dis & Swin-T &
                DeiT-S & DeiT-S-dis & Swin-T \\
                \midrule
                
                0.3048 & 0.7156 & \bf 0.8593  &
                0.5952 & 0.8857 & \bf 0.8345 &
                0.8138 & 0.9983 & 0.8030 &
                0.8638 & \bf 0.9896 & 0.6541 \\
                
                
                
                \bottomrule
        \end{tabular}
        \end{small}
        }
        \end{minipage}
        \vskip -0.1in
        \caption{\small Qualitative Cross-Testing Results. Partially occluded images were generated with iGOS++ from the model with bolded number, then tested on multiple networks and we show the prediction confidences on the ground truth class on each network. 
        (Best viewed in color) }
       \vskip -0.2in 
       \label{figure_igos_example}
\end{figure*}

\begin{figure}[htb]
\centering
\includegraphics[width=0.76\linewidth]{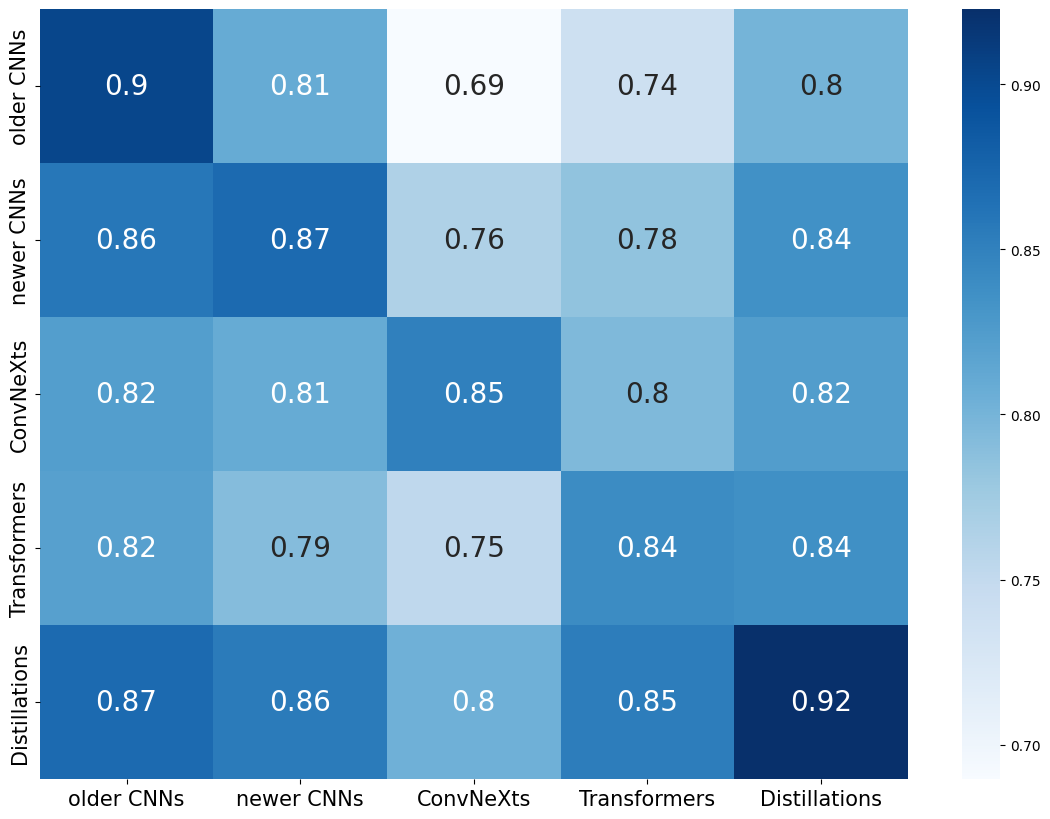}
\vskip -0.13in
\caption{\small Confusion matrix among model groups. The rows are the models used to generate the attribution maps and the columns are the models that the attribution map is cross-tested on. The diagonal values reflect intra-group differences}
 \vskip -0.2in 
\label{5_confusion}
\end{figure}

We also show the average similarity among the different clusters in the confusion matrix in Fig.~\ref{5_confusion}. It can be seen that the confusion matrix is not symmetric -- if we generate the heatmap from old CNNs, the insertion scores among all types of networks are consistently high. However if we generate the heatmap with ConvNeXts, then the insertion score into old CNNs are significantly lower. This shows that older CNNs are more singularly minded and utilized fewer features that are more likely a subset of what is used in newer networks. Newer networks have relatively similar cross-testing insertion scores around 0.8. However, distilled transformers have more similarity with both newer CNNs and other transformers. The conclusion is that these model families still sometimes use different features to classify, which points to potential benefits from ensembles with a model in each cluster. A quick test showed that a simple average of the prediction of ConvNeXt-T (82.1\% accuracy on ImageNet), Swin-T (81.2\%) and ResNet50-c2 (80.0\%) would achieve 82.9\% accuracy, surpassing all individual models.

Figure~\ref{figure_igos_example} shows qualitative results from cross-testing. One can again  see that distilled transformer models sometimes obtain high confidence with a few regions shown. In the \texttt{Bakery} image in the second column, the heatmap is generated with Swin-T, but DeiT-S-distilled have higher confidence than Swin-T, showing that they required less information to obtain more confident predictions. On the other hand, ConvNeXt-T and the ResNets have lower confidence, showing that they may be using different features not shown in this occluded image. In the \texttt{Spoonbill} image in the 3rd column generated by VGG, most other networks were also able to obtain a confident classification. Yet, VGG fares quite poorly on masked images  generated by other networks, showing their overreliance on  specific features that may not be present under those masks.

\vspace{-0.05in}
\section{Conclusion}\vspace{-0.03in}
In this paper, we proposed two novel methodologies, sub-explanation counting and cross-testing, that utilize deep explanation algorithms to collect dataset-wide statistics for understanding the decision-making behaviors of different visual recognition backbones. 
Our analysis indicates that different types of visual recognition models exhibit quite different behaviors along the concept axes of disjunctivism and compositionality. Among other findings, one finding of note is that the choice of normalization strongly affect the compositionality of the model. Receptive field size and data augmentation were shown to also affect model behavior. With cross-testing we  characterized the feature-use landscape of  model families. We hope the insights from our studies could help people better understand decision-making mechanisms of deep visual models and inspire thoughts about future model designs. 
\vspace{-0.1in}
\subsubsection*{Acknowledgements}
This work is partially supported by NSF-1751402 and ONR/NAVSEA contract N00024-10-D-6318.
We also like to thank Dr. Tom Dietterich for his valuable suggestions and Lianghui Wang for assisting with HPC resources.

{
    \small
    \bibliographystyle{ieeenat_fullname}
    \bibliography{egbib}
}












\clearpage
\appendix
\noindent                                                                                                               \textbf{\huge Appendix}
\vspace{0.25in}

\section{Information of the models used in our experiments}

\begin{table}[tbh]
	\centering
	\resizebox{1\linewidth}{!}{%
	\begin{tabular}{lccccc}
		\toprule
		\bf Model  & \bf VGG19 & \bf ResNet50 & \bf ResNet50-C1 & \bf ResNet50-C2 & \bf ResNet50-D\\
		\midrule
		Params & 144M & 25M & 25M & 25M & 25M\\
		Top-1 acc & 74.5 & 76.1 & 79.8 & 80.0 & 79.8\\
  \midrule
        \bf Model  & \bf ConvNeXt-T & \bf Swin-T & \bf Nest-T & \bf DeiT-S & \bf PiT-S\\
        \midrule
        Params & 25M & 28M & 28M & 22M & 23M\\
        Top-1 acc & 82.1 & 81.2 & 81.5 & 79.9 & 80.9 \\
        \midrule
         \bf Model & \bf DeiT-S-distilled & \bf PiT-S-distilled & \bf LeViT-256 & & \\
         \midrule
         Params & 22M & 23M & 19M & & \\
         Top-1 acc  & 81.2 & 81.9 & 81.6 & & \\
		\bottomrule
	\end{tabular}
	}
 \vskip -0.1in
 \caption{Information of the models used in our experiments along with the number of learnable parameters and the Top-1 accuracy on ImageNet-1K.}
    \label{tab:model_info}
 \vskip -0.1in
\end{table}

\begin{table}[tbh]
	\centering
 
	\resizebox{1.0\linewidth}{!}{%
	\begin{tabular}{lccccc}
		\toprule
		\bf Model & \bf ResNet50-A2 & \bf Swin-T & \bf DeiT-S & \bf DeiT-S-distilled & \bf LeViT-256 \\
		\midrule
		Params & 25M & 28M & 22M & 22M & 19M \\
		Top-1 acc & 79.73 $\pm$ 0.15 & 81.09 $\pm$ 0.05 & 79.72 $\pm$ 0.07 & 80.94 $\pm$ 0.16 & 78.76 $\pm$ 0.04 \\
		\bottomrule
	\end{tabular}
	}
 \vskip -0.1in
 \caption{The number of learnable parameters and the Top-1 accuracy on ImageNet-1K of the models we trained. The left side of the symbol $\pm$ is mean value and the right side is standard deviation}
    \label{tab:model_trained_info}
    \vskip -0.1in
\end{table}

\begin{table}[tbh]
	\centering
 
	\resizebox{1.0\linewidth}{!}{%
	\begin{tabular}{lcccc}
		\toprule
		\bf Model & \bf ConvNeXt-T & \bf ConvNeXt-T-3 & \bf ConvNeXt-T-3-GN & \bf ConvNeXt-T-3-BN \\
		\midrule
		Top-1 acc & 82.1 & 81.3 & 82.0 & 80.8 \\
        \midrule
		\bf Model & \bf Swin-T & \bf Swin-T-4 & \bf Swin-T-4-GN & \bf Swin-T-4-BN \\
		\midrule
		Top-1 acc & 81.2 & 81.1 & 81.1 & 80.6 \\
		\bottomrule
	\end{tabular}
	}
 \vskip -0.1in
 \caption{The number of the Top-1 accuracy on ImageNet-1K of the models we trained and the original models.}
    \label{tab:model_trained_info_2}
    \vskip -0.1in
\end{table}

In Table~\ref{tab:model_info} we give some information of the models used in our experiments. In Table~\ref{tab:model_trained_info} and ~\ref{tab:model_trained_info_2} we gave some information about the models we trained ourselves. Note that most of the models have similar sizes to remove a potential confounding factor of the analysis. As the Windows of Swin Transformer must not overlap, during the training of the Swin-T with a small receptive field, we opted to modify the window size to 4 only for the first two stages instead of applying a uniform change to 3 for all stages.

\section{More Results on Minimal Sufficient Explanations and Their Sub-Explanations}

\subsection{Complexity of the algorithms for computing subexplanations.}
The runtimes (in seconds) for identifying all sub-explanations for a single image with a single NVIDIA Tesla V100 GPU are 30 (ResNet50) and 765 (Swin-T), averaged over the first 5,000 images from the ImageNet validation set. Swin-T is much slower due to the huge amount of subexplanations. Note that this runtime pertains only to this analysis to uncover insights, and is not critical for any realistic inference task. There is no additional spatial complexity beyond running inference on the networks.

\subsection{More Information on activation map values}

\begin{figure}[htb]
\centering
\includegraphics[width=0.9\linewidth]{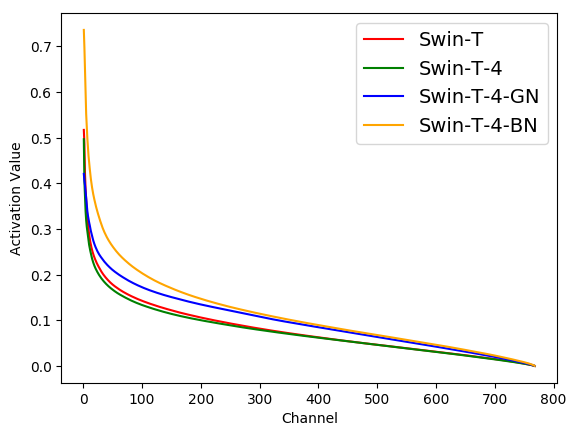}
\vskip -0.1in 
\caption{\small Sorted average values of the maximal activation in each image for each channel in the last block for Swin-T variants}
\label{activation_swin}
\end{figure}

Fig.~\ref{activation_swin} shows the activation map values on Swin Transformers (the main paper has it on ConvNeXt) and indeed the trend is clear that the top feature channels when using BN are much more dominant than with GN and LN as well.

\subsection{Random Seeds Experiment for Establishing Statistical Significance}
In Sec. 4.1, we show the count of MSEs and subexplanations among different networks. To check the statistical significance of our results, we used several representative models trained by ourselves with different random seeds to count the number of MSEs and subexplanations. The results are provided in the Table~\ref{tab:sub_explain_seeds}. These results confirm that the main results are statistically significant: 
1) newer CNN model (ResNet50-A2) tends to have smaller MSE sizes. 
2) Transformer model Swin-T and DeiT-S without distillation have significantly more subexplanations than CNNs and transformers with distillation; 
3) Transformers after distillation have more MSEs with less subexplanations: DeiT-S-dis has significantly more and smaller MSEs than DeiT-S.

\begin{table*}[tbh]
\centering
\vspace{-0.1in}
\resizebox{0.99\linewidth}{!}{%
\begin{small}
\begin{tabular}{lccccccc}
		\toprule
		\multicolumn{2}{c}{\bf Model} & \multicolumn{2}{c}{\bf MSEs} & \multicolumn{4}{c}{\bf Number of Subexplanations}\\
		
		\bf Type & \bf Name & \bf Count &  \bf Size & \bf $\geq$ 80\% & \bf $\geq$ 70\% & \bf $\geq$ 60\% & \bf $\geq$ 50\% \\
        \midrule
		newer CNNs &ResNet50-A2 & 9.10 $\pm$ 0.33 & 5.99 $\pm$ 0.13 & 53.11 $\pm$ 26.93 & 84.60 $\pm$ 37.19 & 138.91 $\pm$ 44.95 & 213.82 $\pm$ 68.34 \\
        \midrule
        Transformers &Swin-T  &  8.51 $\pm$ 0.24 & \textbf{8.20} $\pm$ 0.07 & \textbf{224.57} $\pm$ 59.89 & \textbf{840.20} $\pm$ 117.28 & \textbf{2615.55} $\pm$ 214.73 & \textbf{6349.22} $\pm$ 488.30 \\
        &DeiT-S & 7.72 $\pm$ 1.25 & 8.01 $\pm$ 0.74 & 127.85 $\pm$ 36.73 & 486.02 $\pm$ 89.53 & 1542.36 $\pm$ 350.51 & 3730.73 $\pm$ 1256.73 \\
        \midrule		
        Distillations &DeiT-S-dis  & 10.55 $\pm$ 0.57 & 5.72 $\pm$ 0.21 & 59.96 $\pm$ 26.74 & 129.87 $\pm$ 36.46 & 242.67 $\pm$ 41.90 & 431.75 $\pm$ 64.78 \\
		&LeViT-256 & \textbf{12.47} $\pm$ 0.56 & 5.49 $\pm$ 0.16 & 50.60 $\pm$ 29.12 & 103.10 $\pm$ 37.72 & 164.53 $\pm$ 46.62 & 231.20 $\pm$ 41.67 \\
\bottomrule
\end{tabular}
\end{small}
}
\vspace{-0.1in}
\caption{Beam search results to locate MSEs of seed experiments. Confidence represents average amount of nodes with a classification confidence higher than that threshold w.r.t. the confidence of the whole image. The results are shown in the form of mean $\pm$ standard deviation obtained with 3 different seeds. Statistical significance were affirmed with T-tests} 
\label{tab:sub_explain_seeds}
\end{table*}

\subsection{Effect of Perturbation Style}

\begin{table*}
\vskip -0.05in
\centering
\resizebox{0.68\linewidth}{!}{%
\begin{small}
\begin{tabular}{lcccccccccc}
		\toprule
		\multicolumn{2}{c}{\bf Model}& \bf Perturbation & \multicolumn{2}{c}{\bf MSEs} & \multicolumn{4}{c}{\bf Number of Subexplanations}\\
		\bf Type & \bf Name & & \bf Count &  \bf Size & \bf $\geq$ 80\% & \bf $\geq$ 70\% & \bf $\geq$ 60\% & \bf $\geq$ 50\% \\ 
        \midrule
		older CNNs & ResNet50 & Blur & 6.76 & 7.28 & 53.68 & 108.55 & 180.44 & 296.92 \\
                 & & Grey & 6.62 & \bf 7.65 & 21.47 & 59.76 & 126.07 & 233.92 \\
        \midrule
        ConvNeXt & ConvNeXt-T & Blur & 10.28 & 6.14 & \bf 980.16 & \bf 2001.67 & \bf 3610.37 & 5360.43\\
                 & & Grey & 9.38 & 6.98 & 81.37 & 235.08 & 463.05 & 769.40 \\
        \midrule
		Transformers & Swin-T & Blur & 8.90 & \bf 8.01 & 221.58 & 882.72 & 2933.03 & \bf 7268.20\\
              & & Grey & 10.03 & 6.99 & \bf171.15 & \bf433.26 & \bf851.81 & \bf1373.90 \\
        \midrule
		Distillations & LeViT-256 & Blur & \bf 12.59 & 5.50 & 54.96 & 103.24 & 177.33 & 253.66 \\
                & & Grey & \bf 11.01 & 6.33 & 42.42 & 96.45 & 168.22 & 234.57 \\
		\bottomrule
\end{tabular}
\end{small}
}
\vspace{-0.1in}
\caption{Results of beam search to locate MSEs and sub-explanations. Confidence represents average amount of nodes with a classification confidence higher than the respective threshold w.r.t. the classification confidence on the whole image}
\label{tab:mse_more}
\vspace{-0.1in}
\end{table*}

In Sec. 4.1, we set the perturbed pixels to a highly blurred version of the original image (hereafter referred to as the \textit{Blur} perturbation style) for the Minimal Sufficient Explanations and Their Sub-Explanations experiments. This is a common approach in model explanation literature to alleviate the adversarialness of perturbations~\cite{2018RISE}. Here we use another method, setting the perturbed pixels to zeros (hereafter referred to as the \textit{Grey} perturbation style), to obtain image perturbations. Noting that these could include additional edges to the image hence distort the predictions. Due to the limitations of the GPU resources, we select one model from each model type, plus one ConvNeXt-T.

Table \ref{tab:mse_more} shows the count of MSEs and their sub-explanations for two different perturbation styles: \textit{Grey} and \textit{Blur}. It can be seen that some of the main  conclusions of the paper still stand when using different types of perturbation: 1) LeViT- 256 (a distilled transformer model) still have a higher mean number of MSEs; 2) Transformers without distillations still have significantly more sub-explanations than other models. 

However, in all models, compared to using the \textit{Blur} perturbation style, the number of sub-explanations decreases significantly when the \textit{Grey} perturbation style is used, which showed that the \textit{Grey} style perturbation is more adversarial than the \textit{Blur} perturbation. Besides, the size of Swin-T MSEs decreased significantly once switched to a \textit{Grey} perturbation, which can be explained by that  the \textit{Grey} perturbation may have had a negative effect on the features of all categories (including other categories that might be confusing), hence decreasing the amount of patches that are needed for transformers to make a confident prediction.

\subsection{More Information on MSEs}

\begin{figure}[tbh]
\centering
\includegraphics[width=0.8\linewidth]{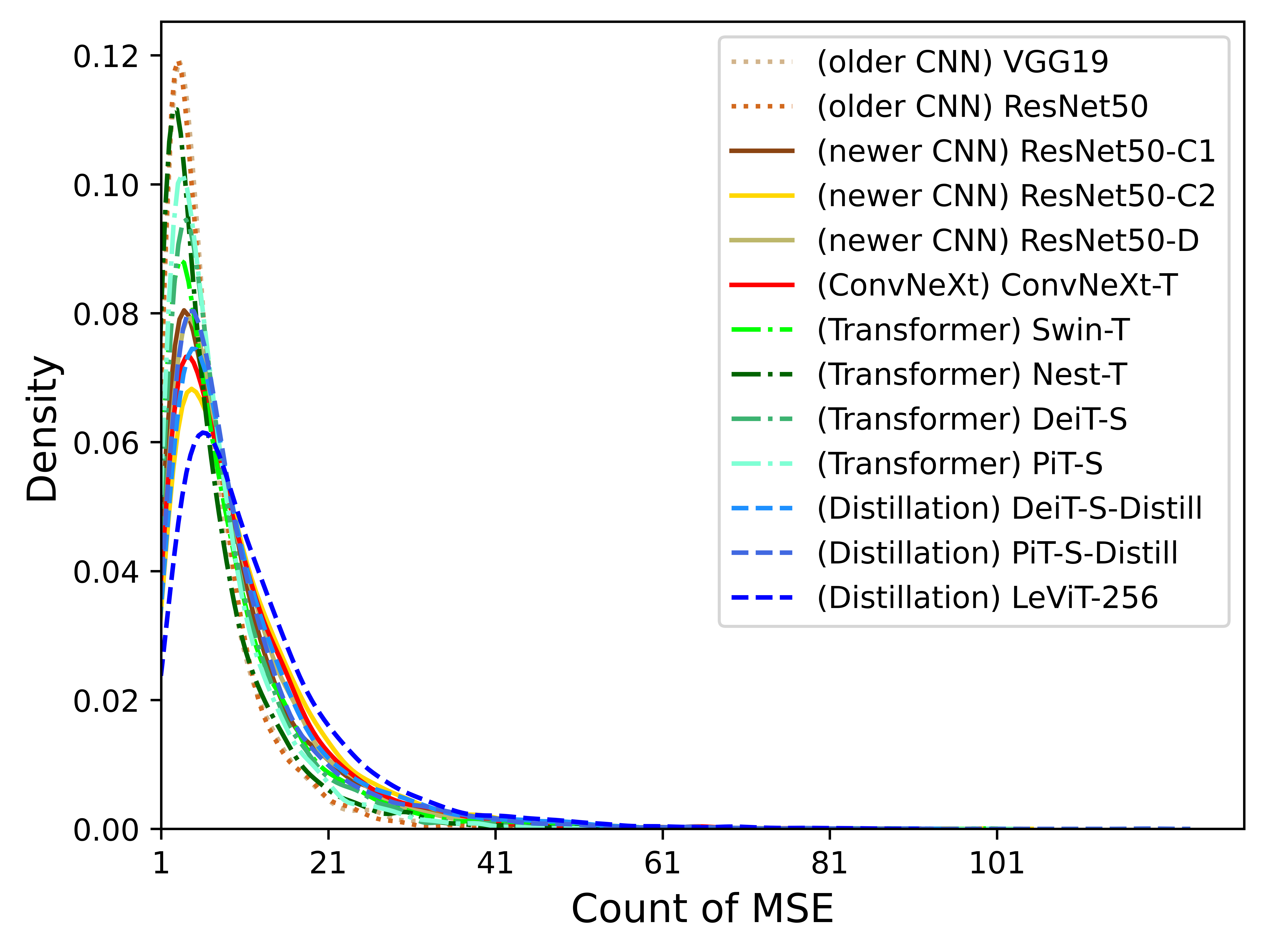}
\vskip -0.1in 
\caption{Distribution of MSE counts on ImageNet images}
\label{figure_mse_number}
\vskip -0.17in 
\end{figure}

\begin{figure}[tbh]
\centering
\includegraphics[width=0.8\linewidth]{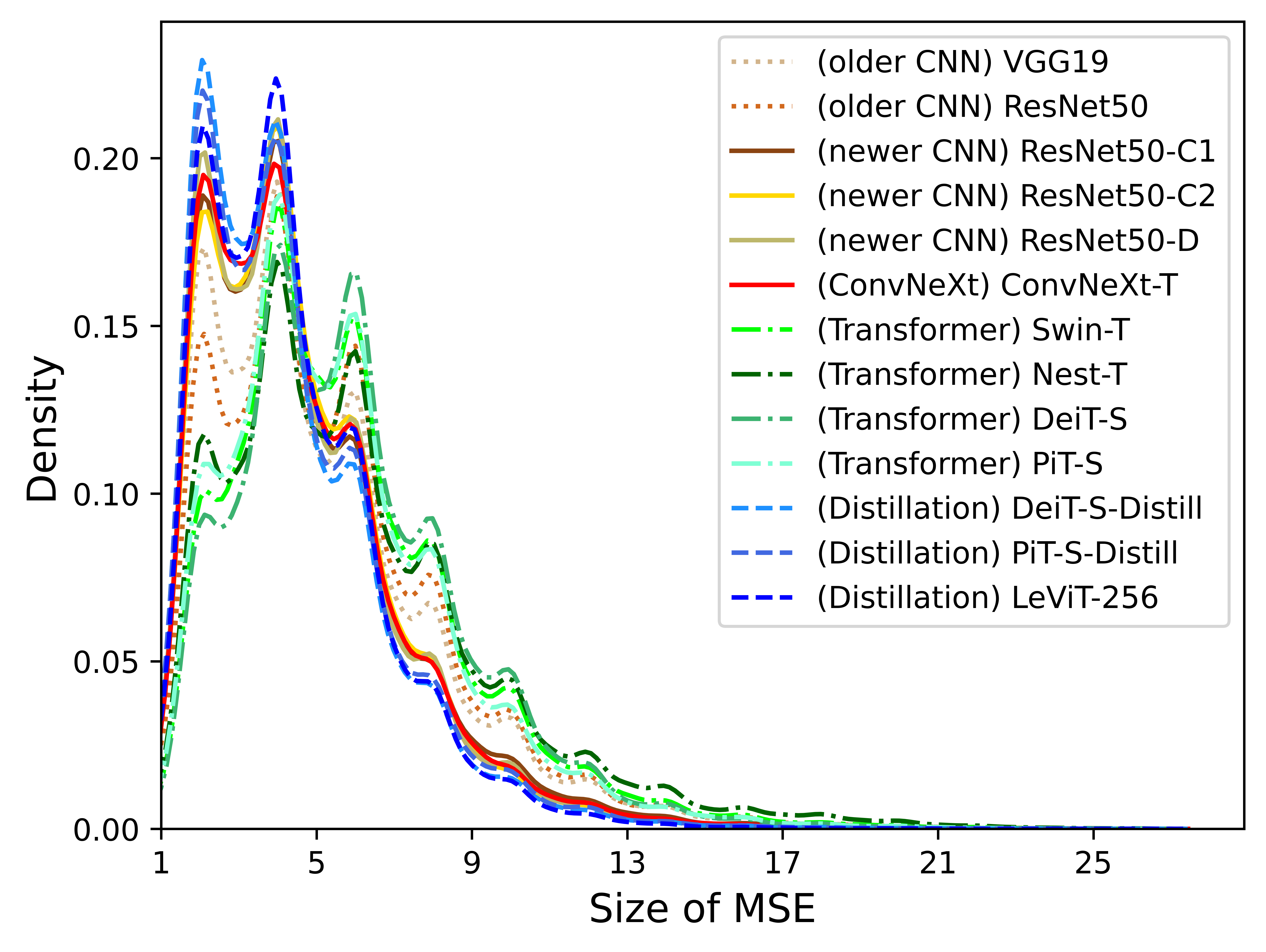}
\vskip -0.1in 
\caption{Distribution of MSE Size on ImageNet images}
\label{figure_mse_size}
\end{figure}

In Fig.~\ref{figure_mse_number} and Fig.~\ref{figure_mse_size} we show the distribution of MSE counts and sizes in individual images. As one can see in Fig.~\ref{figure_mse_number}, in many images, most methods have a small amount of MSEs and similar MSE sizes. However, the peak density is different among different approaches and the tail size is different among different appraoches. older CNNs (VGG19 and ResNet50) and transformers (Swin-T, Nest-T, Deit-S and PiT-S) have the highest density of images with low number of MSEs. In Fig.~\ref{figure_mse_size}, we see that newer CNNs and distilled transformers can classify up to $20\%$ images confidently with as little as \textbf{$2$ patches}, whereas for (non-distilled) transformer models only about $10\%$ of images are explained with only $2$ patches. On the other hand, there are a significant number of images where non-distilled transformers have more patches in their MSEs, as compared to newer CNN models and distilled transformers.

\begin{figure}[htp]
\begin{center}
\includegraphics[width=0.8\linewidth]{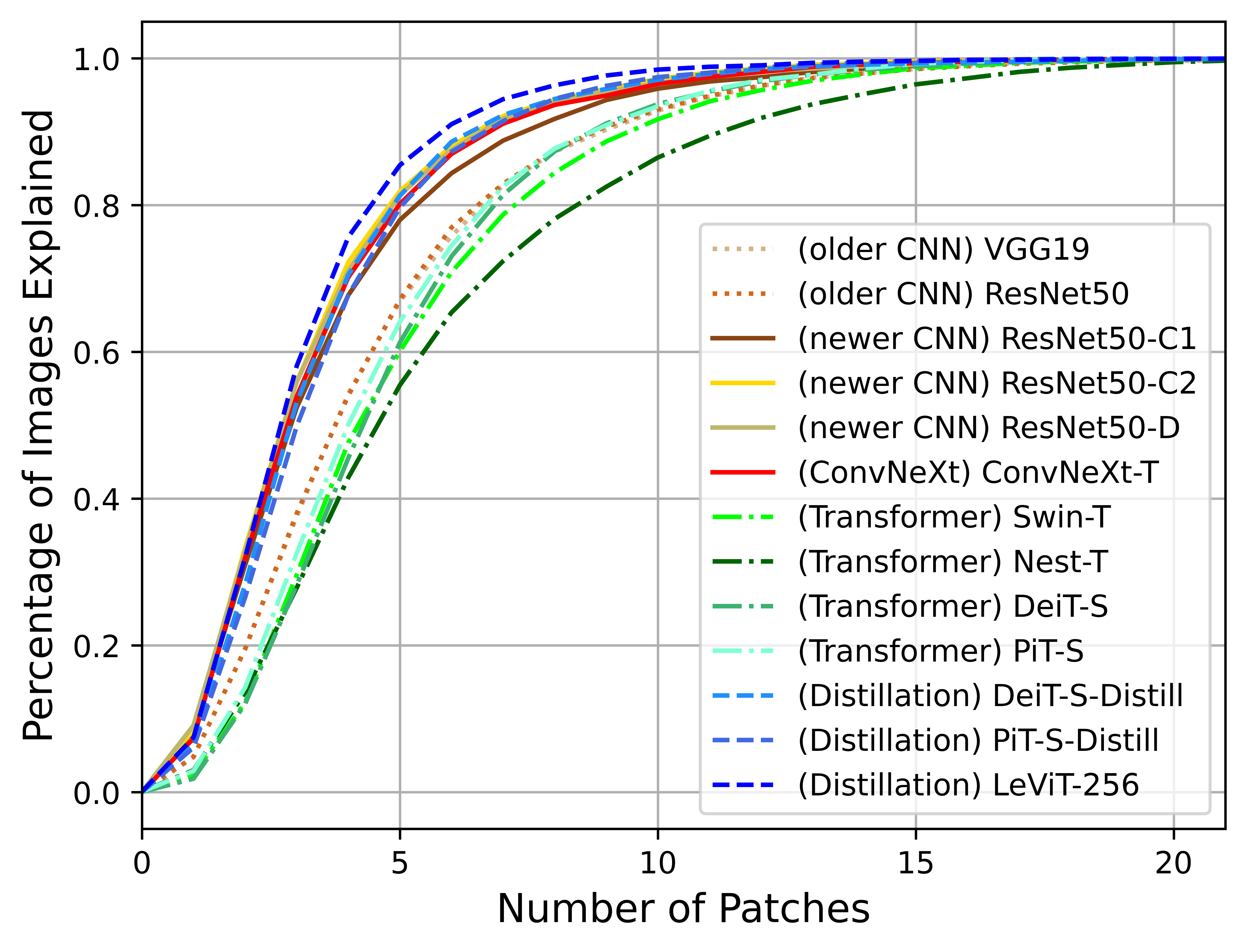}
\end{center}
\vskip -0.2in
\caption{Percentage of images explained by different number of patches.}
 \vskip -0.2in 
\label{figure_percent_explan}
\end{figure}

\begin{figure}[tbh]
        \center
        \begin{minipage}{1.0\linewidth}
        \center
        \setlength\tabcolsep{0.5pt}
        \begin{tabular}{@{}cc@{}}
                Water Tower &  Spoonbill \\
                \includegraphics[width=0.49\linewidth,height=3cm]{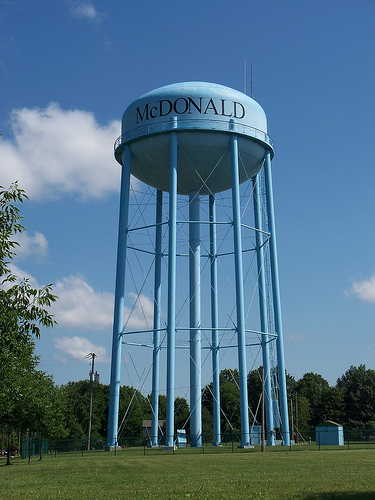} &    \includegraphics[width=0.49\linewidth,height=3cm]{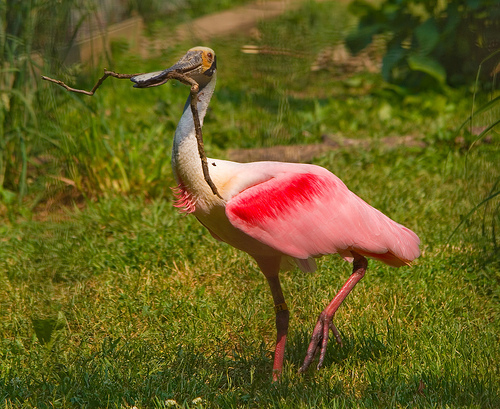}  \\

                \includegraphics[width=0.49\linewidth,height=3cm]{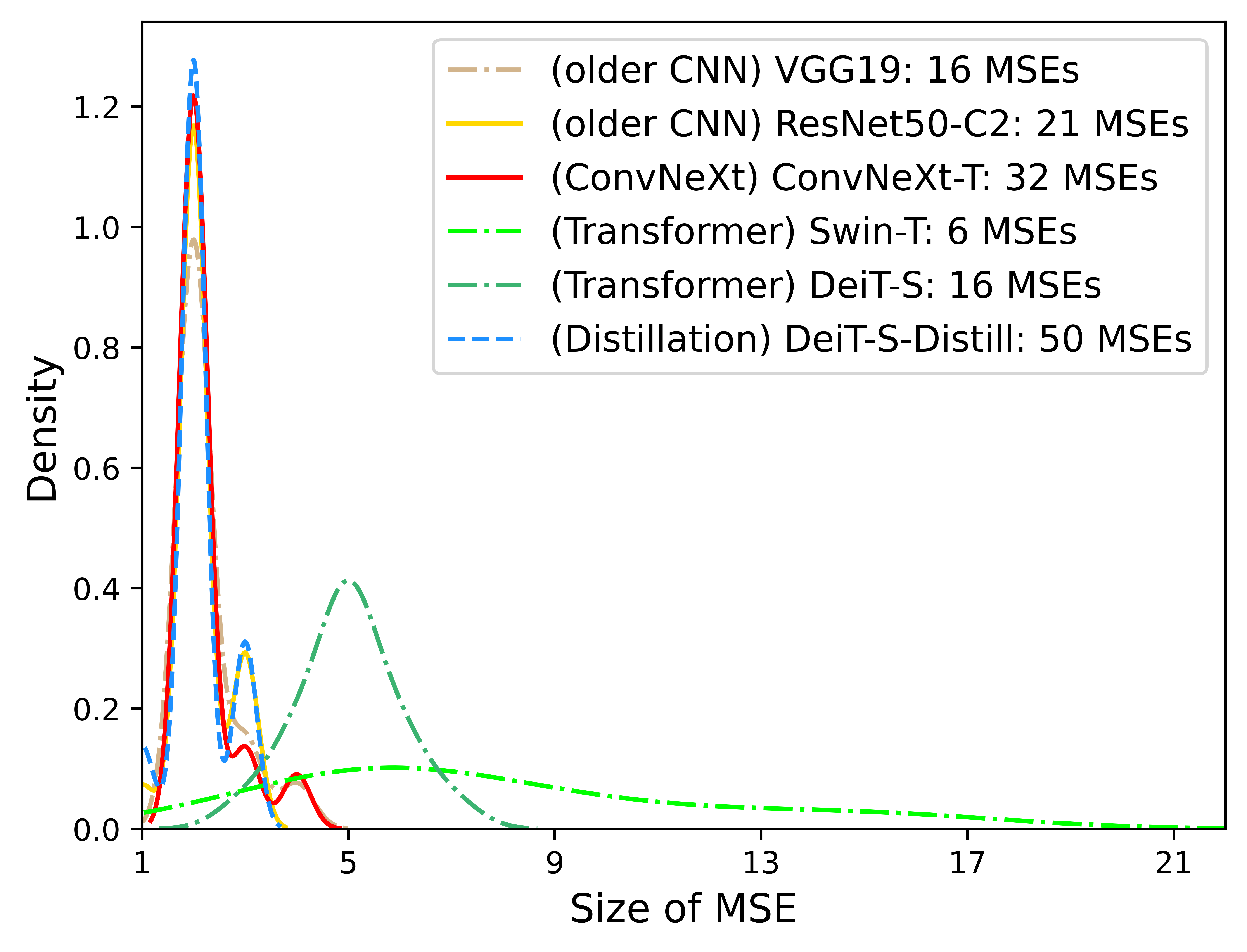} &
                \includegraphics[width=0.49\linewidth,height=3cm]{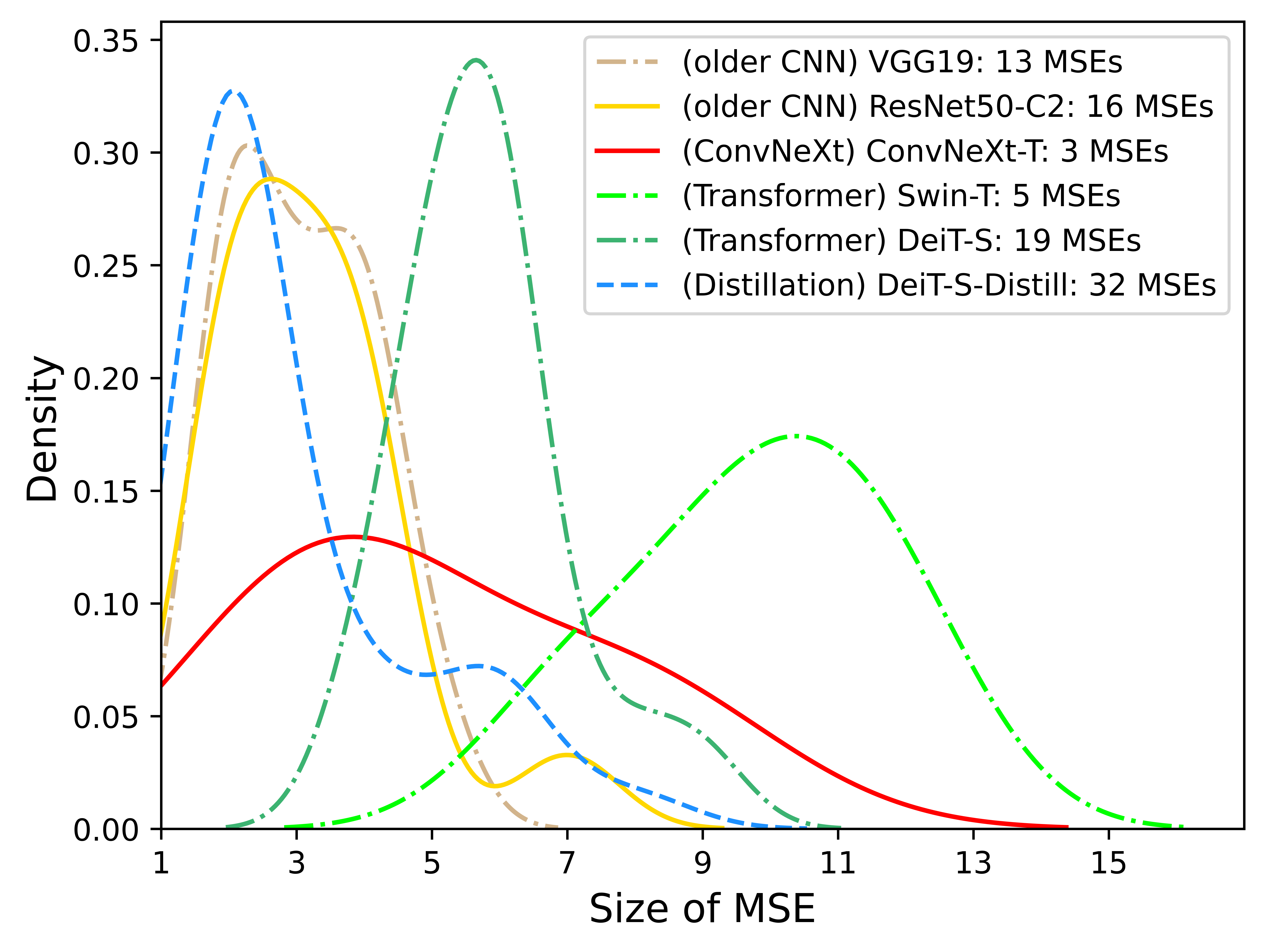} \\
                
                School Bus &  Yurt  \\
               \includegraphics[width=0.49\linewidth,height=3cm]{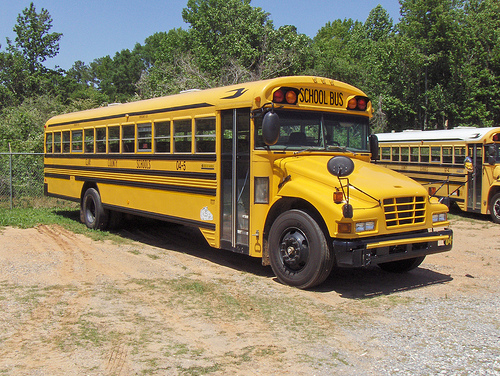} &
                \includegraphics[width=0.49\linewidth,height=3cm]{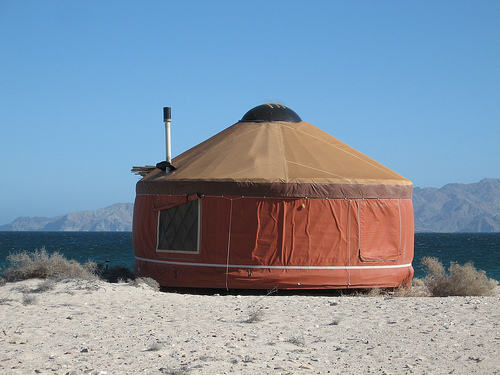} 
                 \\

               \includegraphics[width=0.49\linewidth,height=3cm]{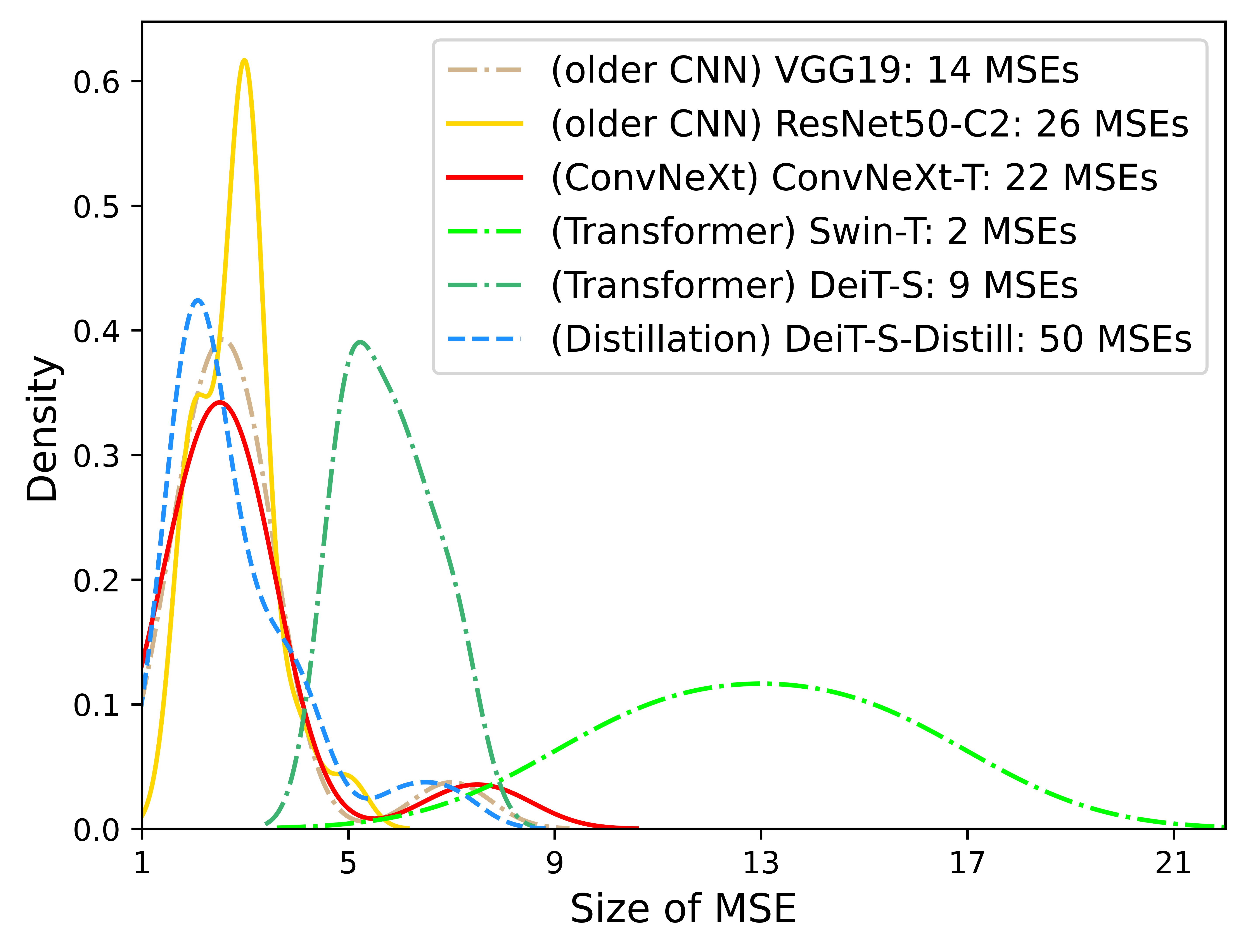} &    \includegraphics[width=0.49\linewidth,height=3cm]{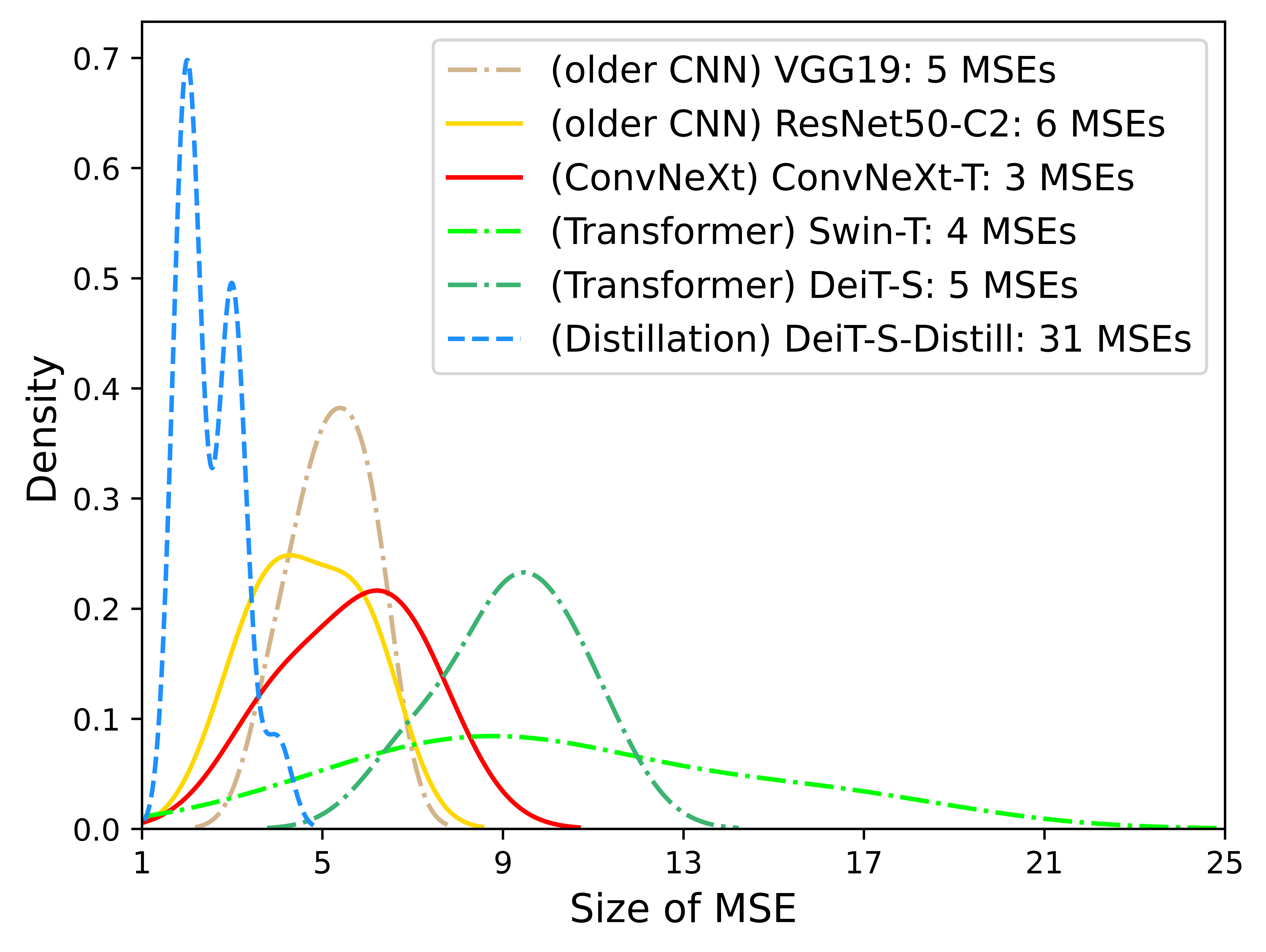} \\

                Black Swan & Pirate Ship \\
                \includegraphics[width=0.49\linewidth,height=3cm]{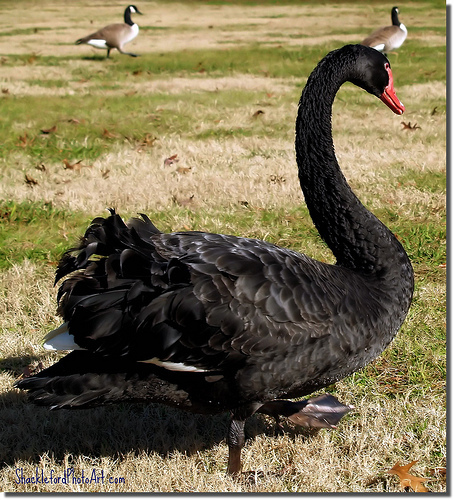} &
                \includegraphics[width=0.49\linewidth,height=3cm]{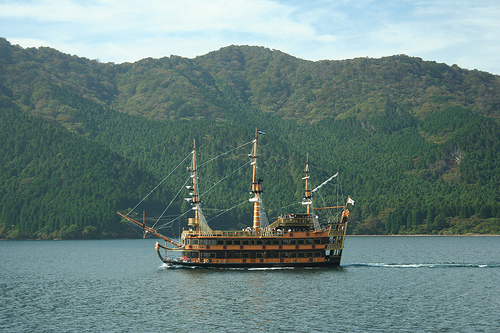} 
                \\
                \includegraphics[width=0.49\linewidth,height=3cm]{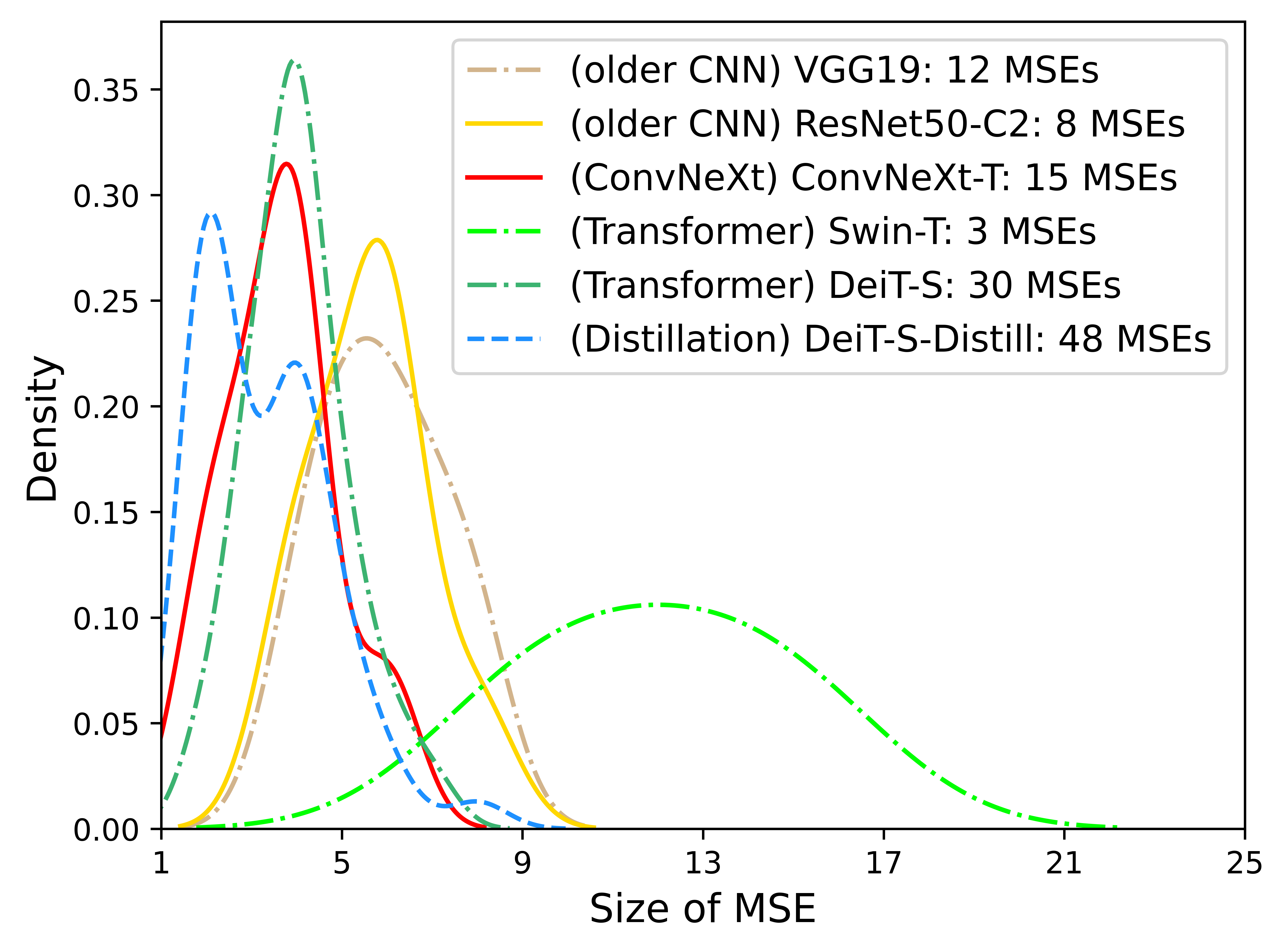} &
                \includegraphics[width=0.49\linewidth,height=3cm]{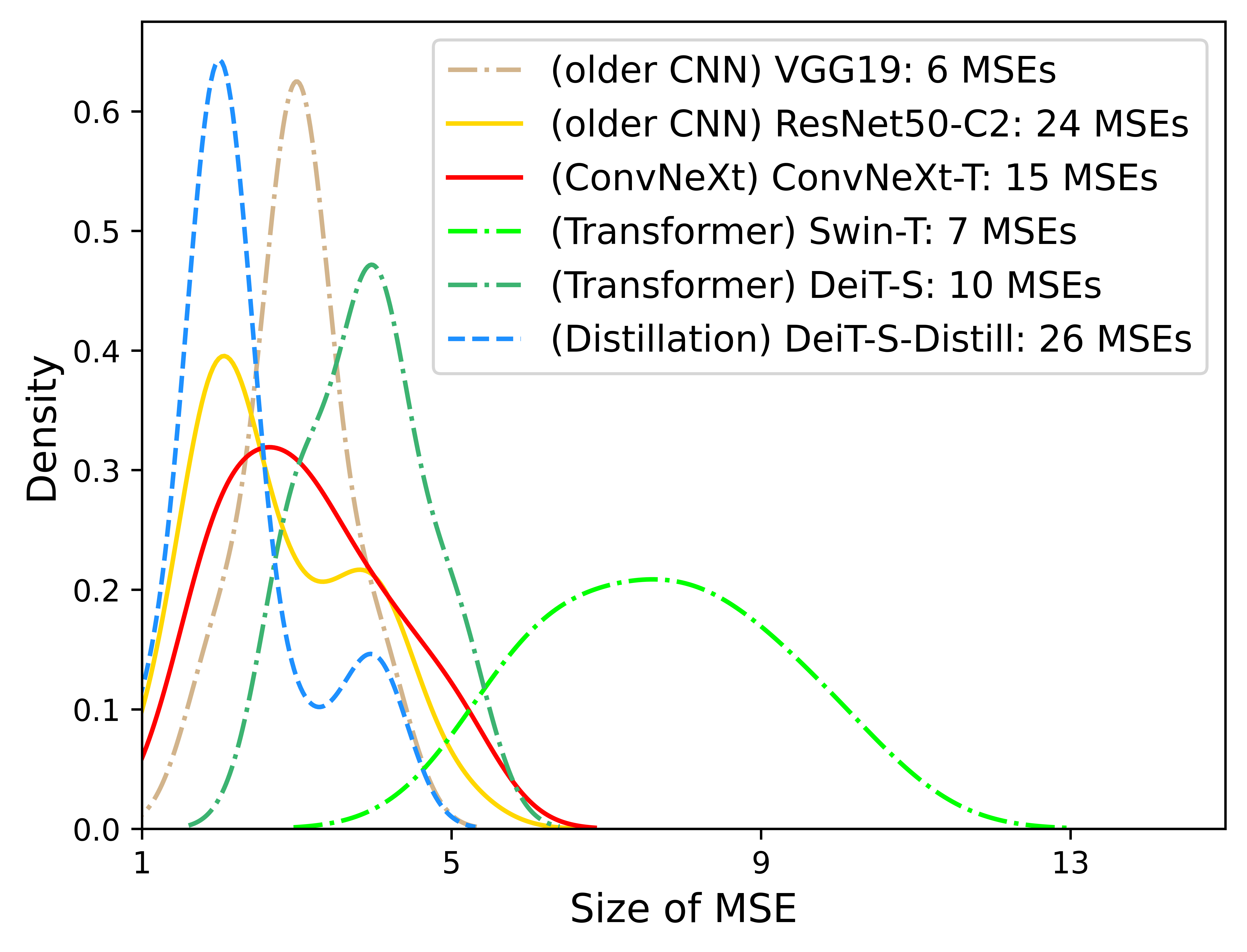} \\

        \end{tabular}
        \vskip -0.15in
        \caption{\small A few example distributions of MSE sizes for different networks on random images. Transformers often have larger MSEs than other networks, and in some images distilled Transformers have significantly more MSEs than other networks}
        \label{figure2}
        \end{minipage}
\end{figure}

Fig.\ref{figure2} shows the distribution of MSE counts (with $\leq 50\%$ overlap) and sizes in a few random images. As one can see, in most images, distilled transformer models often have a significantly higher amount of MSEs. For example, DeiT-S-distilled has $50$ MSEs in the \texttt{Water Tower} and \texttt{School bus} image, much higher than other models that have $3-19$ MSEs. Besides, Transformers (Swin-T and DeiT-S) have significantly higher MSE sizes in these two images than all other methods. 
These plots show a more complete picture of what can happen in each individual image.

We follow \cite{shitole2021one} to plot the percentage of images that can be explained with a small amount of patches. For each number of patches $n$, we plot the total proportion of images that contain at least one MSE with size $\leq$ $n$. For Fig.~\ref{figure_percent_explan}, we use the 5,000 images from ImageNet validation dataset,
we found that the transformers without distillation needed the most amount of patches to explain an image, also include older CNN models. And distilled transformers and newer CNN models in general can explain more images with a small amount of patches.

\begin{table*}[tbh]
	\centering
    \tabcolsep=0.1cm
	\resizebox{0.99\textwidth}{!}{%
	\begin{tabular}{@{ }l|@{ }cc@{ }|@{ }cc@{ }|@{ }cc@{ }|@{ }cc@{ }|@{ }cc@{ }|@{ }cc@{ }|@{ }cc@{ }|@{ }cc@{ }}
		\toprule
		\bf Generation $\rightarrow$ & \multicolumn{2}{c}{\bf VGG19} & \multicolumn{2}{c}{\bf ResNet50} &  \multicolumn{2}{c}{\bf ConvNeXt-T} & \multicolumn{2}{c}{\bf Swin-T} & \multicolumn{2}{c}{\bf Nest-T} & \multicolumn{2}{c}{\bf DeiT-S} & \multicolumn{2}{c}{\bf DeiT-S-distill} & \multicolumn{2}{c}{\bf LeViT-256}\\
		\bf Evaluation $\downarrow$ & Del. & Ins. & Del. & Ins. & Del. & Ins. & Del. & Ins. & Del. & Ins. & Del. & Ins. & Del. & Ins. & Del. & Ins.\\
		\midrule
		\bf VGG19 & 0.139 & 0.742 & 0.196 & 0.732 & 0.336 & 0.517 & 0.195 & 0.673 & 0.178 & 0.568 & 0.200 & 0.608 & 0.187 & 0.634 & 0.280 & 0.626 \\
		\bf ResNet50 & 0.172 & 0.767 & 0.179 & 0.752 & 0.356 & 0.542 & 0.205 & 0.691 & 0.186 & 0.600 & 0.216 & 0.631 & 0.198 & 0.660 & 0.296 & 0.648 \\
		\bf ConvNeXt-T & 0.298 & 0.752 & 0.317 & 0.748 & 0.412 & 0.632 & 0.298 & 0.723 & 0.275 & 0.675 & 0.298 & 0.684 & 0.277 & 0.693 & 0.401 & 0.698 \\
		\bf Swin-T & 0.310 & 0.706 & 0.324 & 0.704 & 0.428 & 0.607 & 0.297 & 0.679 & 0.302 & 0.680 & 0.304 & 0.670 & 0.288 & 0.679 & 0.392 & 0.647 \\
		\bf Nest-T & 0.300 & 0.721 & 0.316 & 0.721 & 0.430 & 0.619 & 0.309 & 0.690 & 0.255 & 0.656 & 0.292 & 0.662 & 0.272 & 0.676 & 0.396 & 0.674 \\
		\bf DeiT-S & 0.261 & 0.758 & 0.284 & 0.749 & 0.404 & 0.602 & 0.277 & 0.713 & 0.258 & 0.657 & 0.244 & 0.664 & 0.246 & 0.689 & 0.365 & 0.665 \\
		\bf DeiT-S-distill & 0.272 & 0.815 & 0.298 & 0.813 & 0.458 & 0.658 & 0.288 & 0.763 & 0.268 & 0.699 & 0.281 & 0.708 & 0.235 & 0.726 & 0.406 & 0.743 \\
		\bf LeViT-256 & 0.320 & 0.844 & 0.346 & 0.844 & 0.516 & 0.703 & 0.336 & 0.800 & 0.309 & 0.741 & 0.334 & 0.744 & 0.305 & 0.767 & 0.433 & 0.781 \\
		\bottomrule
	\end{tabular}
	}
 \vskip -0.1in
    \caption{Score-CAM Cross-Testing. Deletion/Insertion metrics when generating heatmaps using the model on the first row and evaluating the heatmaps by using the model on first column.}
    \label{tab:cross_score}
\end{table*}

Finally, we show more visual examples of Structural Attention Graph (SAG) trees on several images with all the tested approaches (Fig.~\ref{figure_490} - Fig.~\ref{figure_1044_vgg}). Even if we only limited to showing $3$ children per parent node, the size of the trees in Swin-T, DeiT-S and ConvNeXt-T are huge, showing strong evidences that the predictions of them are built by simultaneously taking into account the contributions of many different parts. On the other hand, DeiT-S-distill, VGG and ResNet have very small trees and can obtain very high confidences with very small amount of parts, which makes them to have very few sub-explanations. This shows the difference between compositional and non-compositional disjunctive models more intuitively.

\section{More Results on Cross-Testing}

\subsection{More Results with Score-CAM as the Attribution Map}

In addition to iGOS++, we also performed Cross-Testing using Score-CAM \cite{wang2020score}. This is a visual explanation method based on class activation mapping. We chose this method because it obtains the weight of each activation map through its forward passing score on target class, not depending on gradients. Hence it is a different type of attribution map than the perturbation-based iGOS++. We also normalize the scores for each model \cite{schulz2020restricting}, and use Kernel PCA to project them to 2 dimensions to better visualize their similarities \cite{kernelpca}. 

\begin{figure}[tbh]
\begin{center}
\vskip -0.1in
\includegraphics[width=0.9\linewidth]{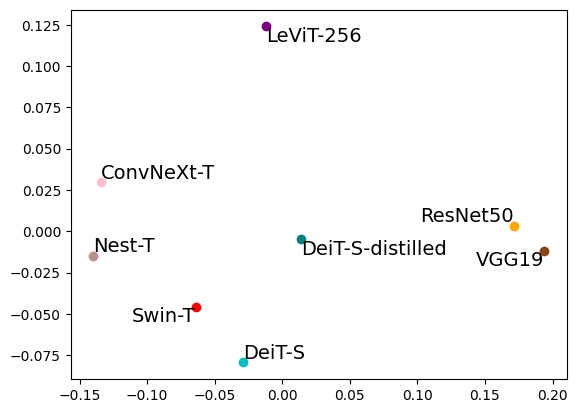}
\end{center}
\vskip -0.3in
\caption{Kernel PCA projections of different models using the insertion metrics obtained with Score-CAM }
\label{figure_pac_score}
\end{figure}

Fig. \ref{figure_pac_score} shows the projection results. We find similar trends as shown in the main paper: that older CNNs (VGG19 and ResNet50) are closer to each other, transformers (Nest-T, Swin-T and DeiT-s) are closer to each other, ConvNeXt-T is closer to transformers, and distillation (from a CNN) brings DeiT-S closer to older CNNs. These four points are consistent with the results we obtained in the main paper with iGOS++. The only difference is that LeViT-256 becomes an outlier. It becomes not very similar with the other distilled model: DeiT-S-distilled.
We do want to note that from Table \ref{tab:cross_score}, we can see that the deletion score related to LeViT-256 is very high. For example, the deletion score of using LeViT-256 to evaluate attribution maps derived from the same algorithm is as high as $0.433$, significantly higher than other methods, which indicates that this method may be less reliable on LeViT-256, in that it highlights some areas that are irrelevant to the prediction. In comparison, the deletion score for iGOS++ on LeViT-256 is only $0.117$. In terms of insertion score, iGOS++ is $0.9 - 0.97$ for all networks if tested on itself, whereas Score-CAM usually averages only around $0.63 - 0.78$, showing that it significantly underperformed iGOS++ by $20\%-30\%$. Note that most attribution maps generate significantly worse deletion scores than I-GOS~\cite{li2021scouter}, the predecessor of iGOS++, and Score-CAM is already an excellent one of its kind, outperforming GradCAM and others based on a third-party benchmark~\cite{li2021scouter}.

\subsection{More Results on iGOS++}

\begin{figure}[htb]
\vskip -0.12in
\centering
\includegraphics[width=0.9\linewidth]{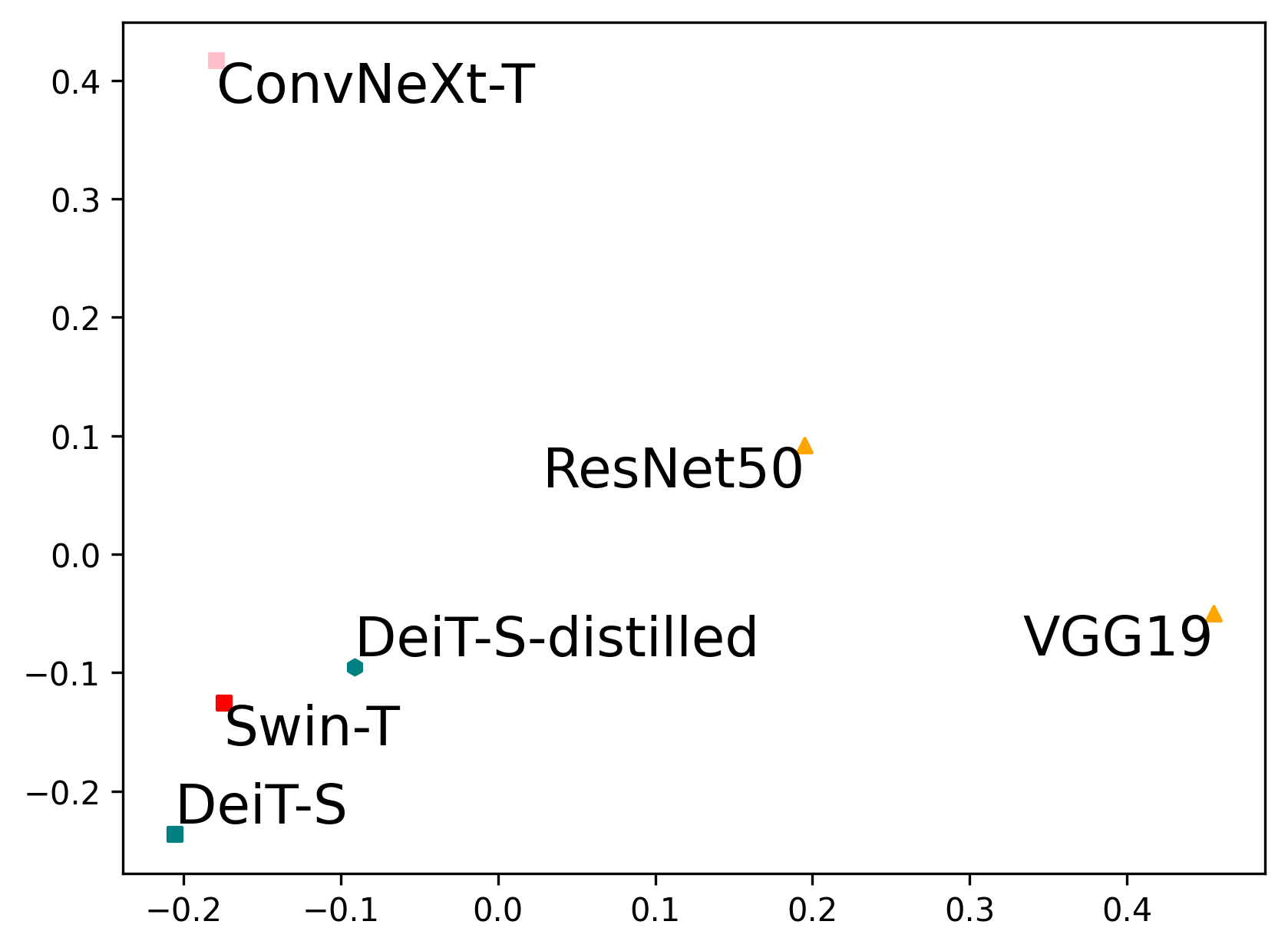}
\vskip -0.15in
\caption{\small Kernel PCA projections of different models using the insertion metrics with a zero-image baseline.} 
\vskip -0.15in
\label{figure_pac_0}
\end{figure}

In Sec. 4.2, we set the perturbed pixels to a highly blurred version of the original image for the cross-testing experiments using iGOS++. Here, we provide cross-testing results for the main models with a zero-image baseline using iGOS++ in Figure~\ref{figure_pac_0}, which showed similar trends as results in the main paper.

\begin{table*}[tbh]
	\centering
    \tabcolsep=0.1cm
	\resizebox{0.8\linewidth}{!}{%
	\begin{tabular}{@{ }l|@{ }cc@{ }|@{ }cc@{ }|@{ }cc@{ }|@{ }cc@{ }|@{ }cc@{ }|@{ }cc@{ }}
		\toprule
		\bf Generation $\rightarrow$ & \multicolumn{2}{c}{\bf VGG19} & \multicolumn{2}{c}{\bf ResNet50} &  \multicolumn{2}{c}{\bf ConvNeXt-T} & \multicolumn{2}{c}{\bf Swin-T} & \multicolumn{2}{c}{\bf DeiT-S} & \multicolumn{2}{c}{\bf DeiT-S-distill} \\
		\bf Evaluation $\downarrow$ & Del. & Ins. & Del. & Ins. & Del. & Ins. & Del. & Ins. & Del. & Ins. & Del. & Ins. \\
		\midrule
		\bf VGG19 & 0.082 & 0.617 & 0.113 & 0.529 & 0.131 & 0.317 & 0.117 & 0.394 & 0.131 & 0.369 & 0.098 & 0.491 \\
		\bf ResNet50 & 0.124 & 0.605 & 0.097 & 0.649 & 0.145 & 0.400 & 0.131 & 0.467 &  0.144 & 0.451 & 0.105 & 0.574 \\
		\bf ConvNeXt-T & 0.203 & 0.709 & 0.204 & 0.709 & 0.155 & 0.683 & 0.182 & 0.621 & 0.200 & 0.603 & 0.150 & 0.703 \\
		\bf Swin-T & 0.224 & 0.672 & 0.226 & 0.665 & 0.224 & 0.553 & 0.155 & 0.745 & 0.223 & 0.598 & 0.166 & 0.682 \\
		\bf DeiT-S & 0.219 & 0.694 & 0.217 & 0.687 & 0.220 & 0.563 & 0.195 & 0.651 & 0.138 & 0.762 & 0.133 & 0.774 \\
		\bf DeiT-S-distill & 0.219 & 0.716 & 0.221 & 0.715 & 0.224 & 0.584 & 0.199 & 0.657 & 0.193 & 0.677 & 0.094 & 0.831 \\
		\bottomrule
	\end{tabular}
	}
 \vskip -0.1in
    \caption{iGOS++ Cross-Testing with a zero-image baseline. Deletion/Insertion metrics when generating heatmaps using the model on the first row and evaluating the heatmaps by using the model on first column.}
    \label{tab:cross_i0}
\end{table*}

\begin{table*}
  \centering
  \resizebox{1\textwidth}{!}{%
   \begin{tabular}{lccccccccccccccccccc}
    \toprule
    \textbf{Generate $\rightarrow$} & \bf VGG19 & \bf ResNet50 & \bf ResNet50-C1 & \bf ResNet50-C2 & \bf ResNet50-D & \bf ConvNeXt-T & \bf ConvNeXt-T-3 & \bf ConvNeXt-T-3-GN & \bf ConvNeXt-T-3-BN & \bf Swin-T & \bf Swin-T-4 & \bf Swin-T-4-GN & \bf Swin-T-4-BN & \bf Nest-T & \bf DeiT-S & \bf PiT-S & \bf DeiT-S-distill & \bf PiT-S-distill & \bf LeViT-256 \\
    \textbf{Evaluate $\downarrow$} & Del & Del & Del & Del & Del & Del & Del & Del & Del & Del & Del & Del & Del & Del & Del & Del & Del & Del & Del \\
    \midrule
    \textbf{VGG19} & 0.944 & 0.863 & 0.786 & 0.778 & 0.807 & 0.651 & 0.651 & 0.653 & 0.66  & 0.715 & 0.717 & 0.732 & 0.765 & 0.647 & 0.699 & 0.707 & 0.789 & 0.8   & 0.796 \\

    \textbf{ResNet50} & 0.88  & 0.926 & 0.824 & 0.821 & 0.839 & 0.719 & 0.724 & 0.72  & 0.739 & 0.767 & 0.768 & 0.781 & 0.821 & 0.705 & 0.758 & 0.767 & 0.844 & 0.85  & 0.851 \\

     \bf ResNet50-C1    & 0.855 & 0.851 & 0.911 & 0.838 & 0.852 & 0.745 & 0.752 & 0.741 & 0.774 & 0.77  & 0.771 & 0.784 & 0.821 & 0.717 & 0.763 & 0.769 & 0.833 & 0.847 & 0.84 \\

    \bf ResNet50-C2    & 0.863 & 0.867 & 0.858 & 0.923 & 0.872 & 0.772 & 0.78  & 0.77  & 0.805 & 0.799 & 0.799 & 0.81  & 0.843 & 0.751 & 0.79  & 0.801 & 0.86  & 0.871 & 0.867 \\

    \bf ResNet50-D     & 0.857 & 0.859 & 0.84  & 0.84  & 0.902 & 0.752 & 0.757 & 0.751 & 0.776 & 0.783 & 0.782 & 0.794 & 0.83  & 0.736 & 0.776 & 0.783 & 0.847 & 0.86  & 0.854 \\

    \bf ConvNeXt-T & 0.823 & 0.821 & 0.803 & 0.803 & 0.814 & 0.947 & 0.797 & 0.797 & 0.801 & 0.796 & 0.792 & 0.805 & 0.824 & 0.758 & 0.783 & 0.795 & 0.824 & 0.835 & 0.821 \\

    \bf ConvNeXt-T-3 & 0.815 & 0.809 & 0.773 & 0.778 & 0.797 & 0.767 & 0.95  & 0.762 & 0.78  & 0.773 & 0.769 & 0.779 & 0.824 & 0.721 & 0.752 & 0.766 & 0.805 & 0.818 & 0.804 \\

    \bf ConvNeXt-T-3-GN & 0.829 & 0.822 & 0.809 & 0.819 & 0.832 & 0.808 & 0.807 & 0.97  & 0.81  & 0.815 & 0.814 & 0.823 & 0.839 & 0.776 & 0.811 & 0.82  & 0.848 & 0.861 & 0.843 \\

    \bf ConvNeXt-T-3-BN & 0.832 & 0.832 & 0.822 & 0.828 & 0.839 & 0.796 & 0.805 & 0.789 & 1.237 & 0.81  & 0.804 & 0.823 & 0.861 & 0.749 & 0.782 & 0.792 & 0.84  & 0.859 & 0.85 \\

    \bf Swin-T & 0.824 & 0.818 & 0.79  & 0.79  & 0.809 & 0.749 & 0.745 & 0.75  & 0.753 & 0.943 & 0.831 & 0.818 & 0.861 & 0.751 & 0.788 & 0.792 & 0.835 & 0.844 & 0.834 \\

    \bf Swin-t-4 & 0.811 & 0.807 & 0.755 & 0.759 & 0.782 & 0.732 & 0.729 & 0.731 & 0.738 & 0.819 & 0.939 & 0.836 & 0.855 & 0.734 & 0.773 & 0.78  & 0.82  & 0.83  & 0.816 \\

    \bf Swin-t-4-GN & 0.804 & 0.804 & 0.767 & 0.765 & 0.787 & 0.747 & 0.742 & 0.75  & 0.757 & 0.82  & 0.828 & 0.954 & 0.857 & 0.749 & 0.79  & 0.794 & 0.831 & 0.841 & 0.825 \\

    \bf Swin-t-4-BN & 0.833 & 0.826 & 0.792 & 0.795 & 0.817 & 0.761 & 0.754 & 0.753 & 0.798 & 0.839 & 0.847 & 0.858 & 1.106 & 0.751 & 0.794 & 0.797 & 0.858 & 0.869 & 0.856 \\

    \bf Nest-T & 0.812 & 0.801 & 0.77  & 0.771 & 0.787 & 0.728 & 0.725 & 0.731 & 0.727 & 0.78  & 0.773 & 0.783 & 0.801 & 0.932 & 0.767 & 0.774 & 0.819 & 0.829 & 0.81 \\

    \bf DeiT-S  & 0.853 & 0.854 & 0.831 & 0.828 & 0.843 & 0.789 & 0.784 & 0.796 & 0.794 & 0.848 & 0.852 & 0.865 & 0.887 & 0.798 & 1.008 & 0.867 & 0.946 & 0.919 & 0.896 \\

    \bf PiT-S & 0.86  & 0.86  & 0.829 & 0.832 & 0.852 & 0.797 & 0.793 & 0.802 & 0.798 & 0.844 & 0.844 & 0.858 & 0.876 & 0.805 & 0.867 & 0.972 & 0.906 & 0.935 & 0.896 \\

    \bf DeiT-S-distill & 0.852 & 0.873 & 0.851 & 0.85  & 0.867 & 0.798 & 0.795 & 0.801 & 0.81  & 0.85  & 0.852 & 0.863 & 0.897 & 0.801 & 0.89  & 0.862 & 0.995 & 0.928 & 0.911 \\

    \bf PiT-S-distill & 0.883 & 0.881 & 0.859 & 0.866 & 0.886 & 0.815 & 0.814 & 0.817 & 0.829 & 0.859 & 0.857 & 0.874 & 0.9   & 0.816 & 0.871 & 0.888 & 0.927 & 0.978 & 0.921 \\

    \bf LeViT-256 & 0.879 & 0.879 & 0.852 & 0.856 & 0.871 & 0.797 & 0.795 & 0.796 & 0.812 & 0.843 & 0.845 & 0.856 & 0.886 & 0.795 & 0.848 & 0.852 & 0.904 & 0.916 & 0.974 \\
    \bottomrule
    \end{tabular}%
    }
    \vskip -0.1in
  \caption{iGOS++ Cross-Testing. Insertion metric when generating heatmaps using the model on the first row and evaluating the heatmaps by using the model on first column.}
    \label{tab:cross_ins}
	\vskip -0.1in
\end{table*}%

\begin{table*}
  \centering
  \resizebox{1\textwidth}{!}{%
    \begin{tabular}{lccccccccccccccccccc}
    \toprule
    \textbf{Generate $\rightarrow$} & \bf VGG19 & \bf ResNet50 & \bf ResNet50-C1 & \bf ResNet50-C2 & \bf ResNet50-D & \bf ConvNeXt-T & \bf ConvNeXt-T-3 & \bf ConvNeXt-T-3-GN & \bf ConvNeXt-T-3-BN & \bf Swin-T & \bf Swin-T-4 & \bf Swin-T-4-GN & \bf Swin-T-4-BN & \bf Nest-T & \bf DeiT-S & \bf PiT-S & \bf DeiT-S-distill & \bf PiT-S-distill & \bf LeViT-256 \\
    \textbf{Evaluate $\downarrow$} & Del & Del & Del & Del & Del & Del & Del & Del & Del & Del & Del & Del & Del & Del & Del & Del & Del & Del & Del \\
    \midrule
    \textbf{VGG19} & 0.111 & 0.16  & 0.175 & 0.178 & 0.172 & 0.191 & 0.201 & 0.194 & 0.196 & 0.173 & 0.172 & 0.17  & 0.154 & 0.198 & 0.128 & 0.189 & 0.145 & 0.144 & 0.15 \\
    
    \textbf{ResNet50} & 0.175 & 0.125 & 0.181 & 0.18  & 0.172 & 0.196 & 0.207 & 0.202 & 0.2   & 0.177 & 0.177 & 0.173 & 0.157 & 0.209 & 0.263 & 0.194 & 0.149 & 0.145 & 0.151 \\
    
    \bf ResNet50-C1    & 0.204 & 0.193 & 0.169 & 0.196 & 0.192 & 0.214 & 0.224 & 0.223 & 0.219 & 0.199 & 0.199 & 0.194 & 0.176 & 0.233 & 0.224 & 0.218 & 0.17  & 0.162 & 0.168 \\
    
    \bf ResNet50-C2    & 0.211 & 0.198 & 0.202 & 0.178 & 0.199 & 0.227 & 0.239 & 0.235 & 0.234 & 0.21  & 0.211 & 0.206 & 0.186 & 0.248 & 0.24  & 0.232 & 0.179 & 0.169 & 0.177 \\
    
    \bf ResNet50-D    & 0.198 & 0.184 & 0.192 & 0.191 & 0.159 & 0.212 & 0.222 & 0.216 & 0.216 & 0.194 & 0.195 & 0.189 & 0.172 & 0.232 & 0.221 & 0.213 & 0.164 & 0.156 & 0.161 \\
    
    \bf ConvNeXt-T & 0.237 & 0.224 & 0.232 & 0.229 & 0.223 & 0.151 & 0.231 & 0.225 & 0.237 & 0.208 & 0.208 & 0.204 & 0.191 & 0.247 & 0.239 & 0.227 & 0.187 & 0.176 & 0.188 \\
    
    \bf ConvNeXt-T-3 & 0.225 & 0.214 & 0.204 & 0.198 & 0.188 & 0.199 & 0.145 & 0.205 & 0.211 & 0.193 & 0.195 & 0.191 & 0.191 & 0.231 & 0.221 & 0.213 & 0.176 & 0.166 & 0.176 \\
    
    \bf ConvNeXt-T-3-GN & 0.238 & 0.225 & 0.222 & 0.217 & 0.202 & 0.215 & 0.229 & 0.151 & 0.233 & 0.206 & 0.209 & 0.201 & 0.189 & 0.246 & 0.236 & 0.227 & 0.183 & 0.172 & 0.184 \\
    
    \bf ConvNeXt-T-3-BN & 0.218 & 0.204 & 0.206 & 0.201 & 0.187 & 0.213 & 0.219 & 0.222 & 0.142 & 0.2   & 0.201 & 0.201 & 0.175 & 0.243 & 0.236 & 0.227 & 0.177 & 0.169 & 0.177 \\
    
    \bf Swin-T & 0.245 & 0.232 & 0.244 & 0.24  & 0.235 & 0.228 & 0.243 & 0.236 & 0.243 & 0.129 & 0.181 & 0.196 & 0.165 & 0.232 & 0.222 & 0.216 & 0.175 & 0.169 & 0.181 \\
    \bf Swin-T-4 & 0.241 & 0.231 & 0.223 & 0.219 & 0.207 & 0.226 & 0.24  & 0.233 & 0.241 & 0.176 & 0.128 & 0.18  & 0.156 & 0.23  & 0.219 & 0.211 & 0.172 & 0.168 & 0.179 \\
    
    \bf Swin-T-4-GN & 0.239 & 0.228 & 0.226 & 0.223 & 0.208 & 0.228 & 0.243 & 0.233 & 0.243 & 0.18  & 0.176 & 0.122 & 0.156 & 0.231 & 0.219 & 0.213 & 0.173 & 0.169 & 0.18 \\
    
    \bf Swin-t-4-BN & 0.233 & 0.225 & 0.222 & 0.218 & 0.204 & 0.231 & 0.245 & 0.239 & 0.239 & 0.185 & 0.18  & 0.176 & 0.109 & 0.238 & 0.226 & 0.225 & 0.173 & 0.17  & 0.179 \\
    
    \bf Nest-T & 0.241 & 0.231 & 0.237 & 0.238 & 0.232 & 0.23  & 0.244 & 0.236 & 0.245 & 0.196 & 0.2   & 0.196 & 0.181 & 0.144 & 0.222 & 0.213 & 0.173 & 0.168 & 0.182 \\
    
    \bf DeiT-S  & 0.244 & 0.231 & 0.249 & 0.249 & 0.239 & 0.247 & 0.266 & 0.255 & 0.266 & 0.204 & 0.202 & 0.197 & 0.18  & 0.241 & 0.127 & 0.214 & 0.144 & 0.165 & 0.178 \\
   
    \bf PiT-S & 0.249 & 0.237 & 0.238 & 0.236 & 0.218 & 0.244 & 0.261 & 0.253 & 0.263 & 0.206 & 0.206 & 0.201 & 0.183 & 0.245 & 0.221 & 0.14  & 0.169 & 0.151 & 0.177 \\
    
    \bf DeiT-S-distill & 0.244 & 0.232 & 0.246 & 0.246 & 0.236 & 0.246 & 0.263 & 0.257 & 0.264 & 0.205 & 0.205 & 0.198 & 0.18  & 0.245 & 0.204 & 0.221 & 0.095 & 0.16  & 0.173 \\
   
    \bf PiT-S-distill & 0.25  & 0.235 & 0.238 & 0.232 & 0.217 & 0.25  & 0.27  & 0.264 & 0.272 & 0.216 & 0.217 & 0.211 & 0.19  & 0.256 & 0.238 & 0.216 & 0.17  & 0.109 & 0.179 \\
    
    \bf LeViT-256 & 0.257 & 0.24  & 0.253 & 0.251 & 0.245 & 0.254 & 0.269 & 0.263 & 0.27  & 0.224 & 0.221 & 0.219 & 0.197 & 0.263 & 0.25  & 0.241 & 0.186 & 0.178 & 0.118 \\
    \bottomrule
    \end{tabular}%
    }
    \vskip -0.1in
  \caption{iGOS++ Cross-Testing. Deletion metric when generating heatmaps using the model on the first row and evaluating the heatmaps by using the model on first column.}
    \label{tab:cross_del}
	\vskip -0.2in
\end{table*}%

In Section 4.2, we provide the the visualized results of different classifiers using Kernel PCA, which show the similarities between different models. Here we show the normalized deletion and insertion scores obtained from different classifiers.We can see that most of the algorithms have similar and fairly low deletion scores and fairly high relative insertion scores. This indicates that the attribution maps found by the algorithm explain the decisions consistently and the model is able to obtain similar confidence as the full image by only using a few top-ranked patches from the attribution map, which proves that the attribution map  algorithm we use is sound as a basis for the cross-testing experiments. Note that the distilled transformer models (DeiT-S-distill, PiT-S-distill and LeViT-256) have slightly higher insertion scores which indicate they need fewer patches to achieve the same confidence as the full image than the other algorithms. 
DeiT-S-distill even has a relative insertion score close to $1$, indicating that in many cases, partially occluded images have been more confidently predicted than the full image. Another observation is that transformers trained with the batch normalization (ConvNeXt-T-3-BN and Swin-T-4-BN) consistently exhibit significantly higher insertion scores, surpassing $1.1$. This finding implies that these transformers require fewer patches to attain higher confidence levels compared to the full image than the other models.

From the results shown in Table~\ref{tab:cross_ins} and ~\ref{tab:cross_del} one can see the significant differences between different models. Cross-testing insertion metric is usually around $80\%$, which indicates that the models agree on less than $90\%$ of the images. The similarity between models of the same category are usually higher, e.g. between VGG19 and ResNet50, and among DeiT-S-distilled, PiT-S-distilled and LeViT-256, also during ResNet50-C1, ResNet50-C2 and ResNet50-D. 
Still, the similarities between Swin-T and other transformer models are higher than between Swin-T and the CNN models.

\begin{figure*}[htp]
        \center
        \begin{minipage}{0.8\linewidth}
        \center
        \setlength\tabcolsep{0.5pt}
        \begin{tabular}{ccc}
        
                Cradle & Cougar & Granny Smith\\
        
                \includegraphics[width=0.33\linewidth,height=3.4cm]{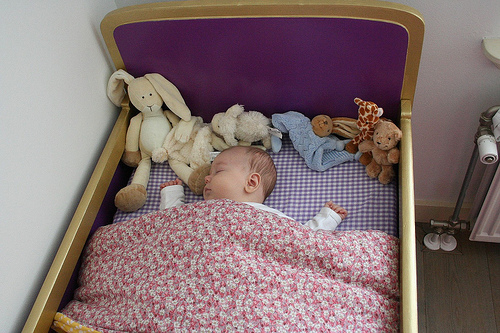} &    \includegraphics[width=0.33\linewidth,height=3.4cm]{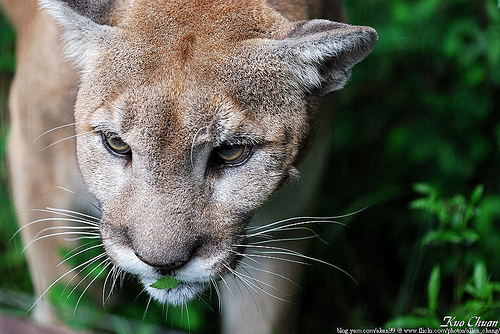} &
                \includegraphics[width=0.33\linewidth,height=3.4cm]{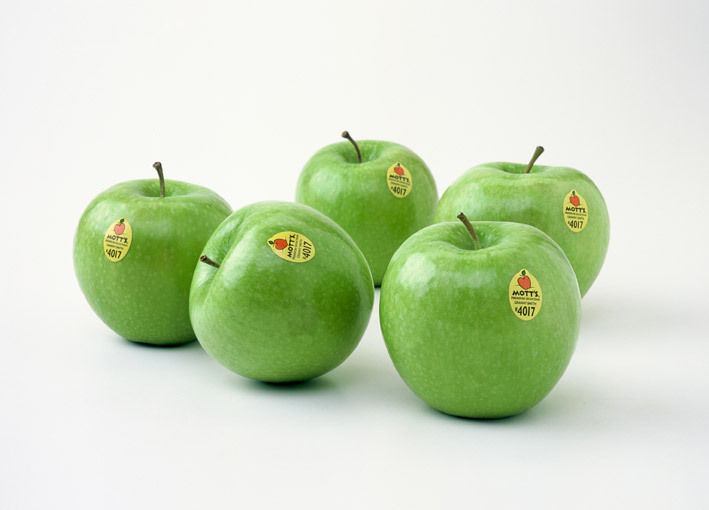} \\

                \includegraphics[width=0.33\linewidth,height=3.4cm]{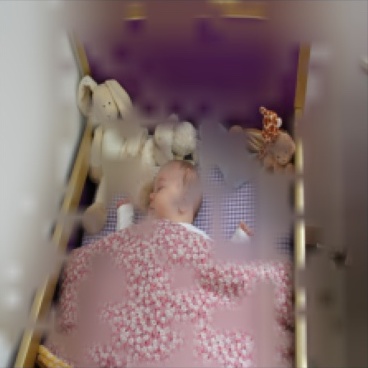} &
                \includegraphics[width=0.33\linewidth,height=3.4cm]{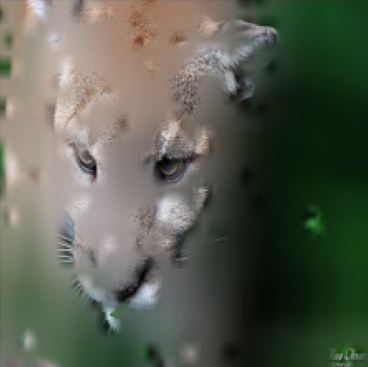} &
                \includegraphics[width=0.33\linewidth,height=3.4cm]{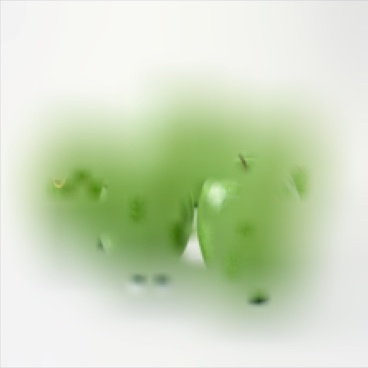}\\
                
        \end{tabular}
        \end{minipage}
        
        \begin{minipage}{0.8\linewidth}
        \center
        \scriptsize
        \resizebox{0.98\linewidth}{!}{%
        \begin{tabular}{@{ }c@{ }|@{ }c@{ }|@{ }c@{ }||@{ }c@{ }|@{ }c@{ }|@{ }c@{ }||@{ }c@{ }|@{ }c@{ }|@{ }c@{ }}
                \multicolumn{9}{c}{Prediction Confidence on the Partially Occluded Image}\\
                \toprule
                
                VGG19 & ResNet50-C2 & ConvNeXt-T &
                VGG19 & ResNet50-C2 & ConvNeXt-T &
                VGG19 & ResNet50-C2 & ConvNeXt-T \\
                \midrule
                0.1646 & 0.1526 & 0.3726 & 
                0.1609 & 0.4226 & 0.2244 &
                0.3474 & 0.0063 & 0.8044 \\
                
                \midrule
                DeiT-S & DeiT-S-dis & Swin-T &
                DeiT-S & DeiT-S-dis & Swin-T &
                DeiT-S & DeiT-S-dis & Swin-T \\
                \midrule
                
                0.1831 & 0.9081 & \textbf{0.9573} &
                0.8468 & 0.9696 & \textbf{0.7739} &
                0.1787 & 0.4536 & \textbf{0.6497} \\

                \bottomrule
        \end{tabular}
        }
        \end{minipage}

        \begin{minipage}{0.8\linewidth}
        \center
        \setlength\tabcolsep{0.5pt}
        \begin{tabular}{ccc}
        
                Shetland sheepdog & Convertible & Mousetrap\\
        
                \includegraphics[width=0.33\linewidth,height=3.4cm]{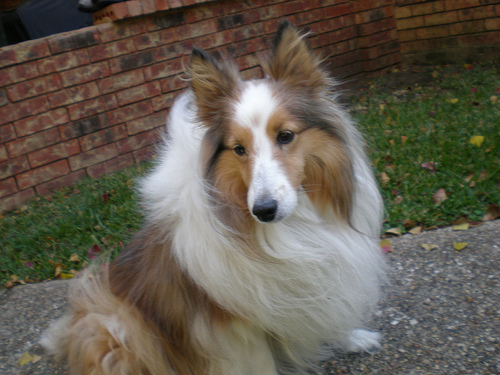} &    \includegraphics[width=0.33\linewidth,height=3.4cm]{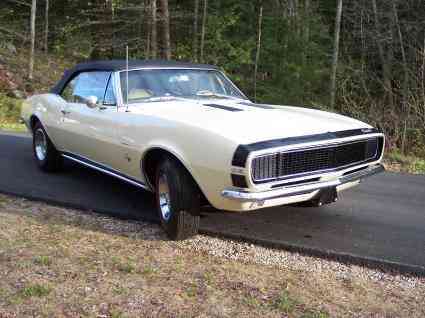} &
                \includegraphics[width=0.33\linewidth,height=3.4cm]{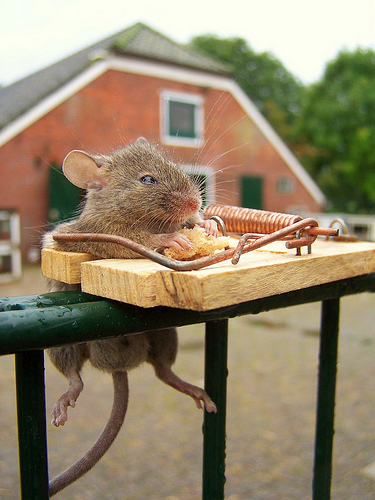} \\

                \includegraphics[width=0.33\linewidth,height=3.4cm]{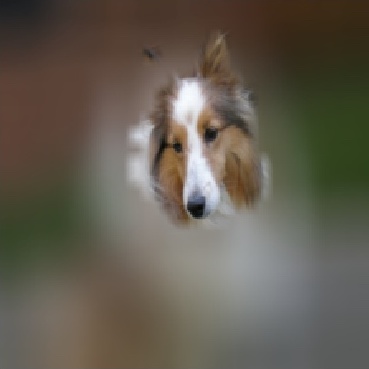} &
                \includegraphics[width=0.33\linewidth,height=3.4cm]{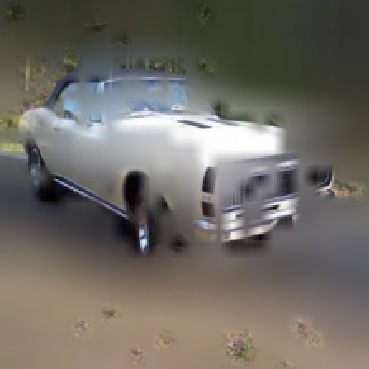} &
                \includegraphics[width=0.33\linewidth,height=3.4cm]{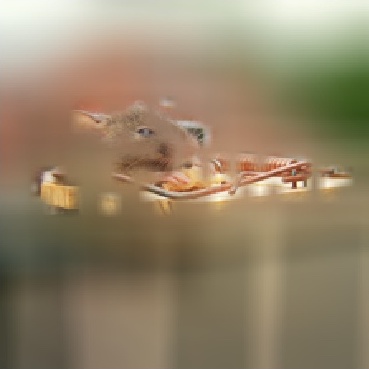}\\
                
        \end{tabular}
        \end{minipage}

        \begin{minipage}{0.8\linewidth}
        \center
        \scriptsize
        \resizebox{0.98\linewidth}{!}{%
        \begin{tabular}{@{ }c@{ }|@{ }c@{ }|@{ }c@{ }||@{ }c@{ }|@{ }c@{ }|@{ }c@{ }||@{ }c@{ }|@{ }c@{ }|@{ }c@{ }}
                \multicolumn{9}{c}{Prediction Confidence on the Partially Occluded Image}\\

                \toprule

                VGG19 & ResNet50-C2 & ConvNeXt-T &
                VGG19 & ResNet50-C2 & ConvNeXt-T &
                VGG19 & ResNet50-C2 & ConvNeXt-T \\
                \midrule
                \bf 0.4311 & 0.6174 & 0.2859 &
                0.1273 & 0.1962  & \bf 0.8120 &
                 0.0026 & 0.1846 & 0.0897 \\

                \midrule
                DeiT-S & DeiT-S-dis & Swin-T &
                DeiT-S & DeiT-S-dis & Swin-T &
                DeiT-S & DeiT-S-dis & Swin-T \\
                \midrule
                
                0.7315 & 0.9677 & 0.8174 &
                0.6469 & 0.8251 & 0.2517 &
                 0.5709 & 0.9832 & \bf 0.8725 \\
                
                \bottomrule
        \end{tabular}
        }
        \end{minipage}
            \vskip -0.1in
        \caption{Qualitative Cross-Testing Results. The partially occluded images were generated using iGOS++ heatmaps on the algorithm with bolded number (not necessarily the highest). Then the same image is tested on multiple algorithms and we show predicted class-conditional probabilities on the ground truth class (written above). }
       \vskip -0.1in \label{figure_igos_example_2}
\end{figure*}

\begin{figure*}[tbh]
	\centering
	\includegraphics[width=0.4\linewidth]{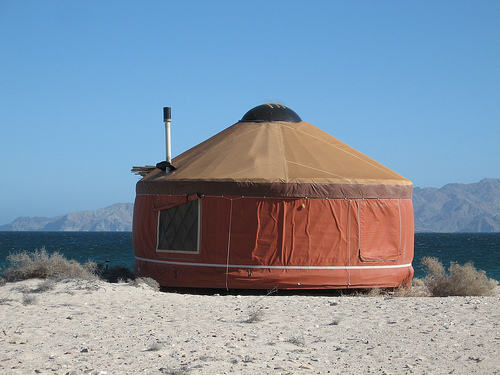}
	\caption{An image of a yurt}
	\label{figure_490}
\end{figure*}

\begin{figure*}[tbh]
    \centering
	\includegraphics[width=0.99\linewidth]{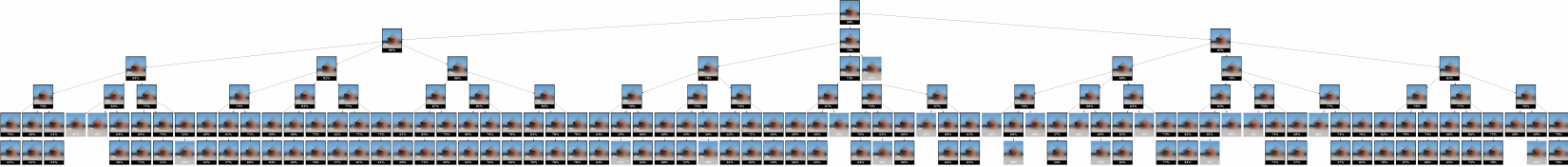}
	\caption{An example SAG tree explaining Swin Transformers on Fig.~\ref{figure_490}. This tree is too big to be visualized efficiently, but the sheer size of it shows the robustness of Swin Transformers to different types of occlusions. It also justifies our approach of looking at statistics rather than the visualization themselves }
	\label{figure_490_swin}
 \end{figure*}

\begin{figure*}[tbh]
	\centering
	\includegraphics[width=0.99\linewidth]{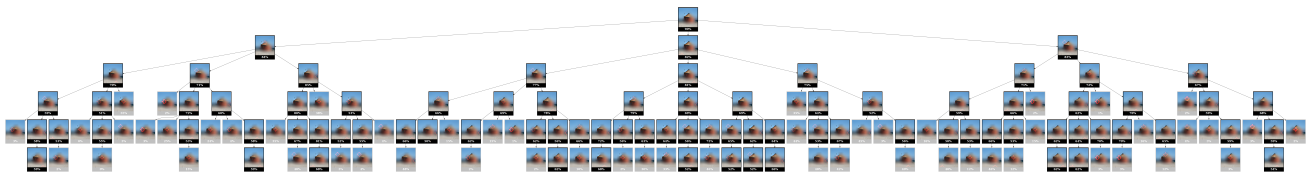}
	\caption{An example SAG tree explaining Fig.~\ref{figure_490} for ConvNeXt-T. The tree size is too large to be visualized properly}
	\label{figure_490_convnext}
\end{figure*}

\begin{figure*}[tbh]
	\centering
	\includegraphics[width=0.99\linewidth]{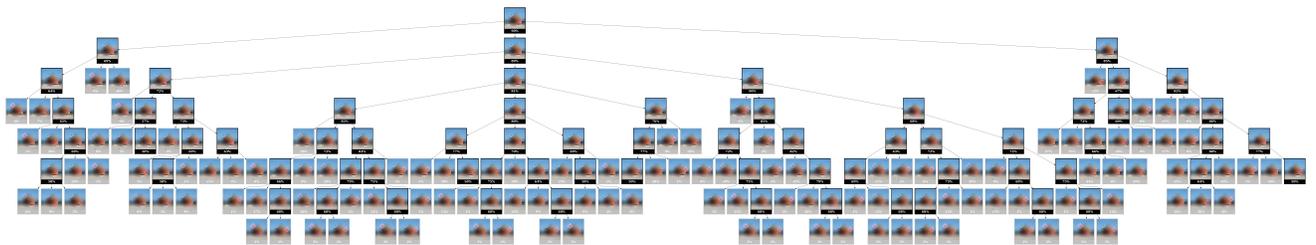}
	\caption{An example SAG tree explaining Fig.~\ref{figure_490} for DeiT-S. The tree size is too large to be visualized properly}
	\label{figure_490_deit_s}
\end{figure*}

\begin{figure*}[tbh]
	\centering
	\includegraphics[width=0.4\linewidth]{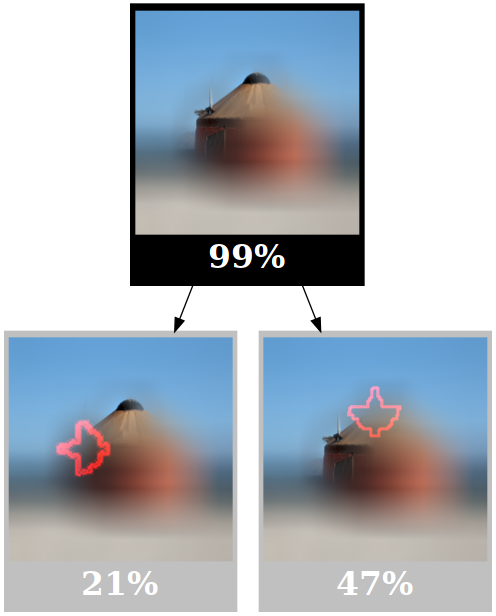}
	\caption{An example SAG tree explaining Fig.~\ref{figure_490} for DeiT-S Distilled}
	\label{figure_490_deit_s_dis}
\end{figure*}

\begin{figure*}[tbh]
	\centering
	\includegraphics[width=0.6\linewidth]{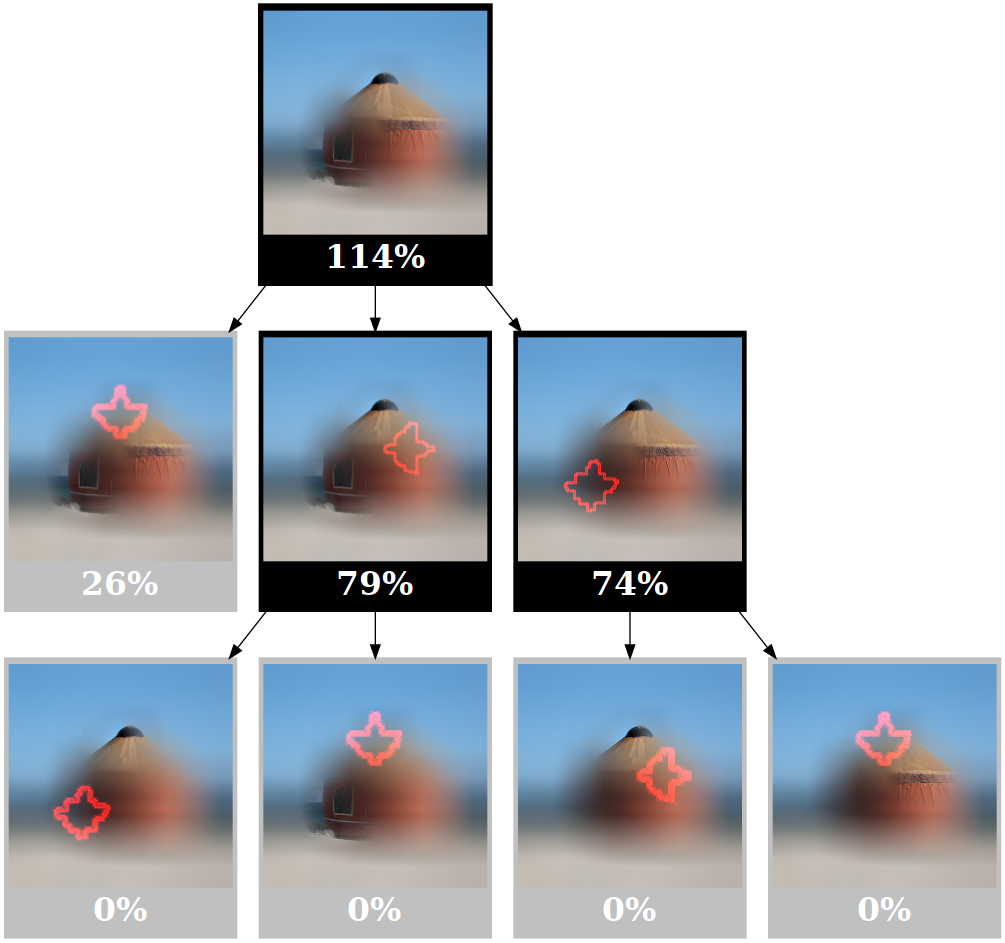}
	\caption{An example SAG tree explaining Fig.~\ref{figure_490} for ResNet-50-C2. It can be seen that the SAG is small and focused on a very specific combination of patches of the sausage}
	\label{figure_490_resnet_c2}
\end{figure*}

\begin{figure*}[tbh]
	\centering
	\includegraphics[width=0.4\linewidth]{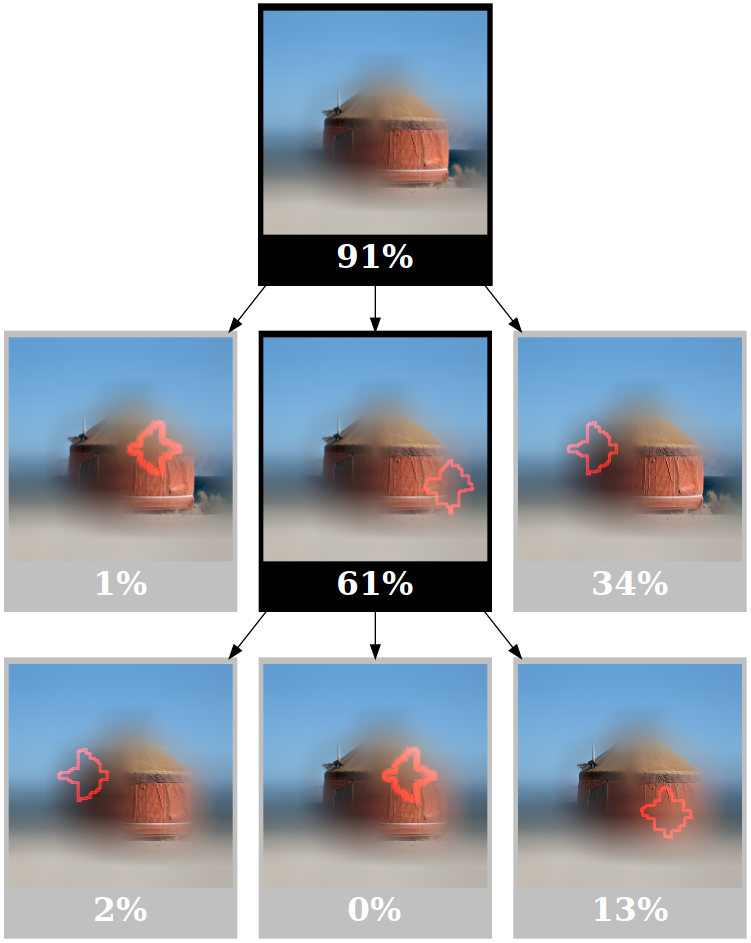}
	\caption{An example SAG tree explaining Fig.~\ref{figure_490} for VGG. It can be seen that in many cases removal of a few parts lead to low-confidence predictions}
	\label{figure_490_vgg}
\end{figure*}

\begin{figure*}
	\centering
	\includegraphics[width=0.55\linewidth]{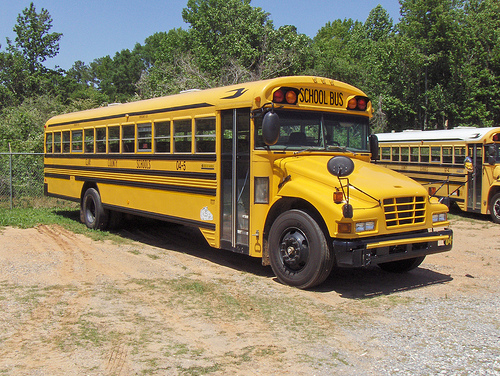}
	\caption{An image of the School Bus}
	\label{figure_1771}
\end{figure*}

\begin{figure*}
\flushleft
	\includegraphics[width=0.99\linewidth]{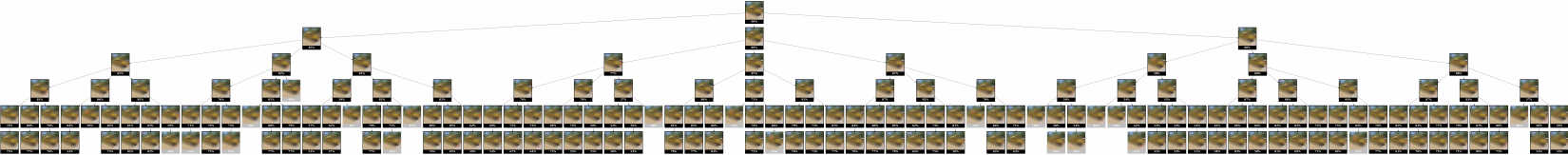}
	\caption{An example SAG tree explaining Swin Transformers on Fig.~\ref{figure_1771}. Again, the tree size is too large to be visualized properly}
	\label{figure_1771_swin}
\end{figure*}

\begin{figure*}
\flushleft
	\includegraphics[width=0.99\linewidth]{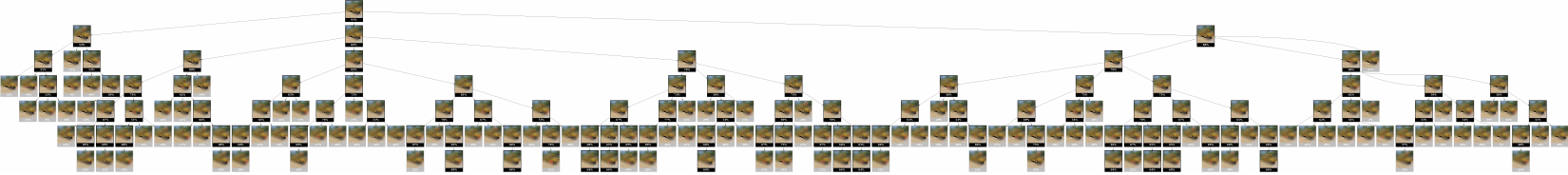}
	\caption{An example SAG tree explaining Fig.~\ref{figure_1771} for ConvNeXt-T. Again, the tree size is too large to be visualized properly}
	\label{figure_1771_convnext}
\end{figure*}

\begin{figure*}
\flushleft
	\includegraphics[width=0.99\linewidth]{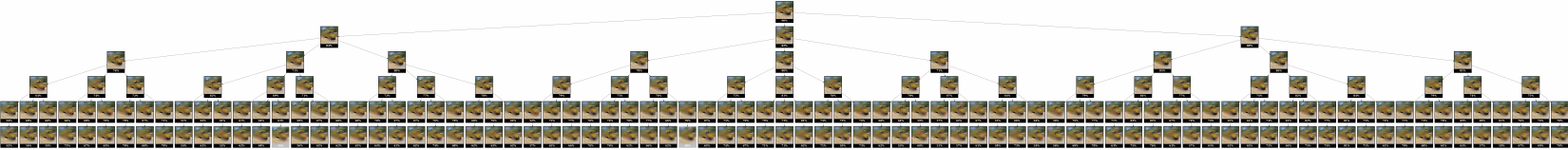}
	\caption{An example SAG tree explaining Fig.~\ref{figure_1771} for DeiT-S. Again, the tree size is too large to be visualized properly}
	\label{figure_1771_deit_s}
\end{figure*}

\begin{figure*}
	\centering
	\includegraphics[width=0.4\linewidth]{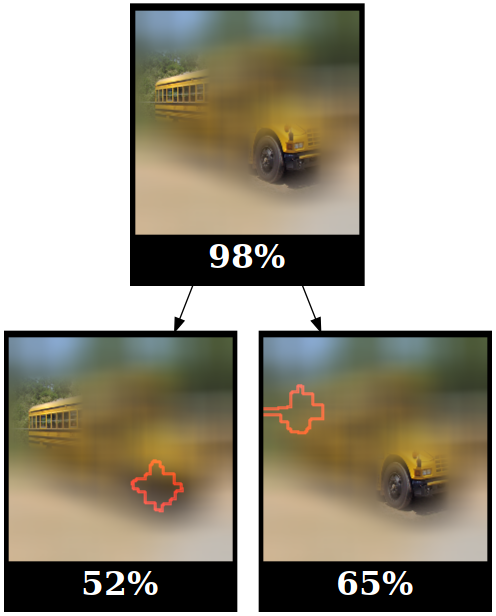}
	\caption{An example SAG tree explaining Fig.~\ref{figure_1771} for DeiT-S Distilled}
	\label{figure_1771_deit_s_dis}
\end{figure*}

\begin{figure*}
	\centering
	\includegraphics[width=0.4\linewidth]{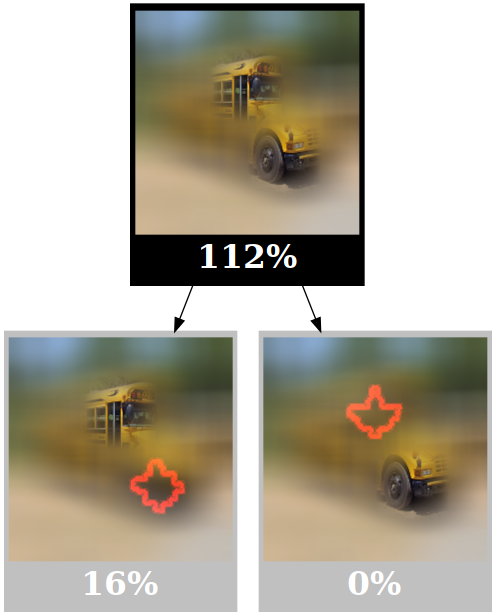}
	\caption{An example SAG tree explaining Fig.~\ref{figure_1771} for ResNet-50-C2.}
	\label{figure_1771_resnet_c2}
\end{figure*}

\begin{figure*}
	\centering
	\includegraphics[width=0.4\linewidth]{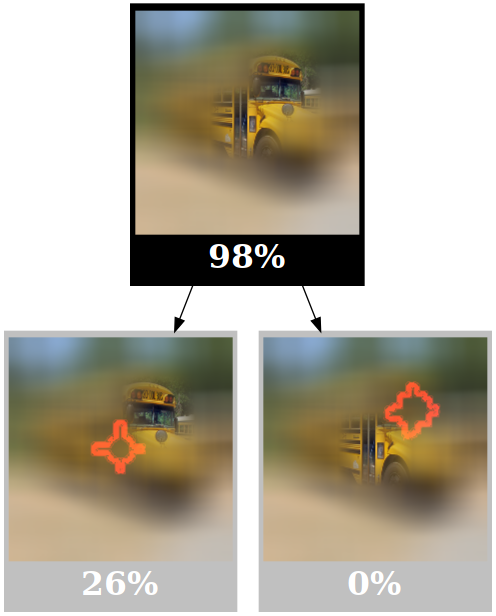}
	\caption{An example SAG tree explaining Fig.~\ref{figure_1771} for VGG.}
	\label{figure_1771_vgg}
\end{figure*}

\begin{figure*}
	\centering
	\includegraphics[width=0.4\linewidth]{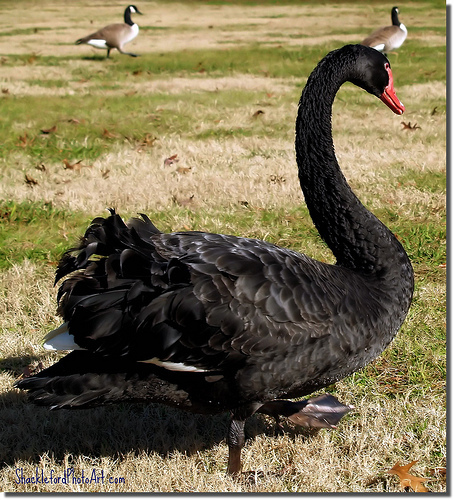}
	\caption{An image of the Black Swan}
	\label{figure_1044}
\end{figure*}

\begin{figure*}
\flushleft
	\includegraphics[width=0.99\linewidth]{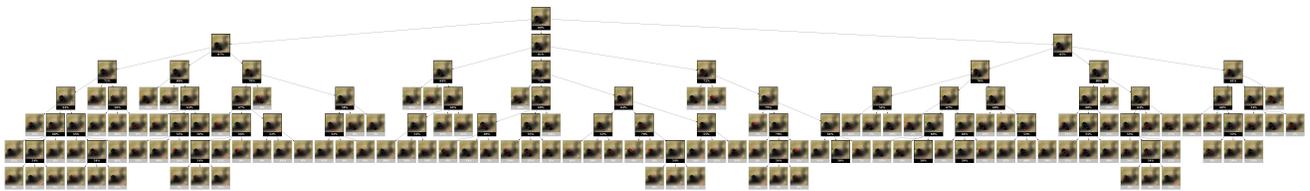}
	\caption{An example SAG tree explaining Swin Transformers on Fig.~\ref{figure_1044}. Again, the tree size is too large to be visualized properly}
	\label{figure_1044_swin}
\end{figure*}

\begin{figure*}
	\centering
	\includegraphics[width=0.99\linewidth]{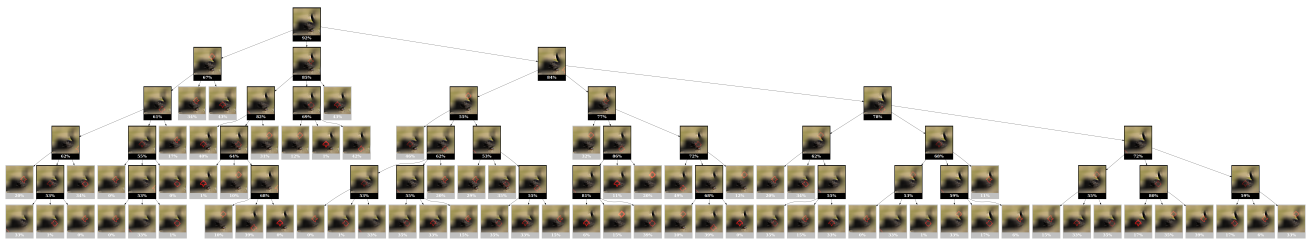}
	\caption{An example SAG tree explaining Fig.~\ref{figure_1044} for ConvNeXt-T. Again, the tree size is too large to be visualized properly}
	\label{figure_1044_convnext}
\end{figure*}

\begin{figure*}
	\centering
	\includegraphics[width=0.99\linewidth]{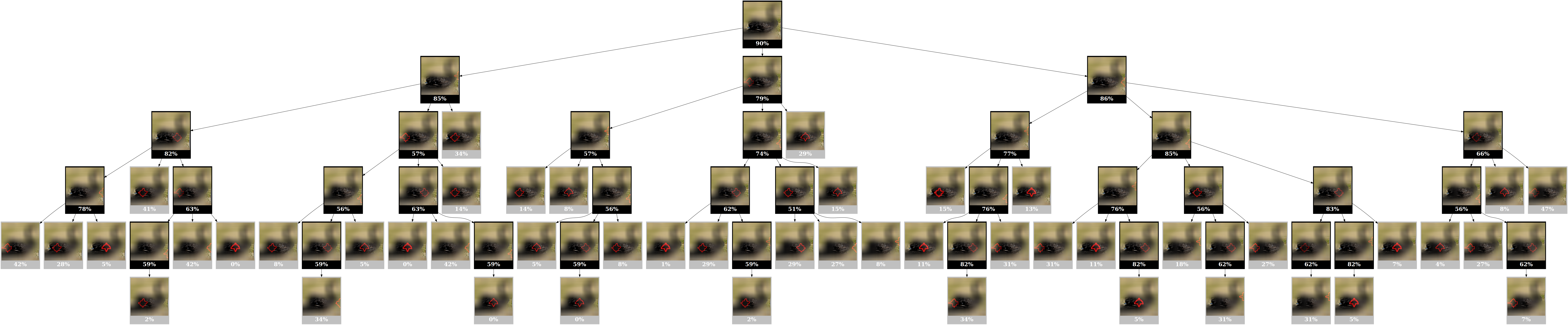}
	\caption{An example SAG tree explaining Fig.~\ref{figure_1044} for DeiT-S. Again, the tree size is too large to be visualized properly}
	\label{figure_1044_deit_s}
\end{figure*}

\begin{figure*}
	\centering
	\includegraphics[width=0.35\linewidth]{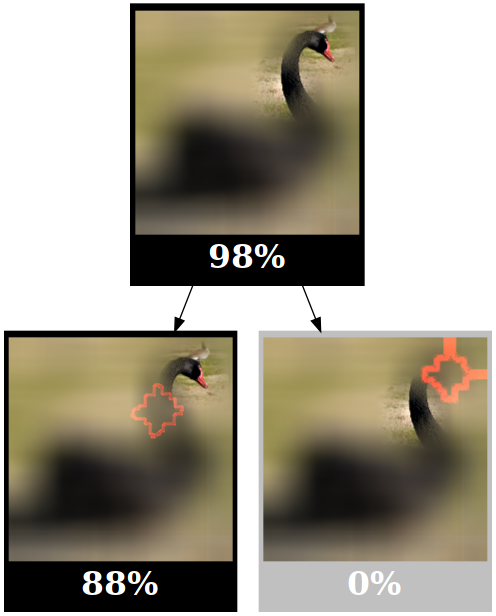}
	\caption{An example SAG tree explaining Fig.~\ref{figure_1044} for DeiT-S Distilled}
	\label{figure_1044_deit_s_dis}
\end{figure*}

\begin{figure*}
	\centering
	\includegraphics[width=0.35\linewidth]{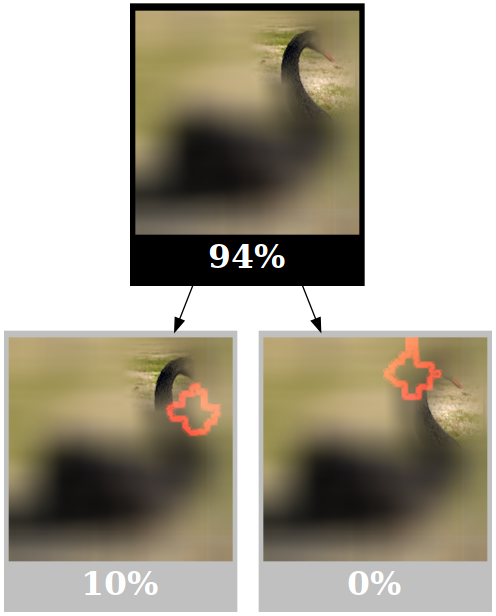}
	\caption{An example SAG tree explaining Fig.~\ref{figure_1044} for ResNet-50-C2.}
	\label{figure_1044_resnet_c2}
\end{figure*}

\begin{figure*}
	\centering
	\includegraphics[width=0.6\linewidth]{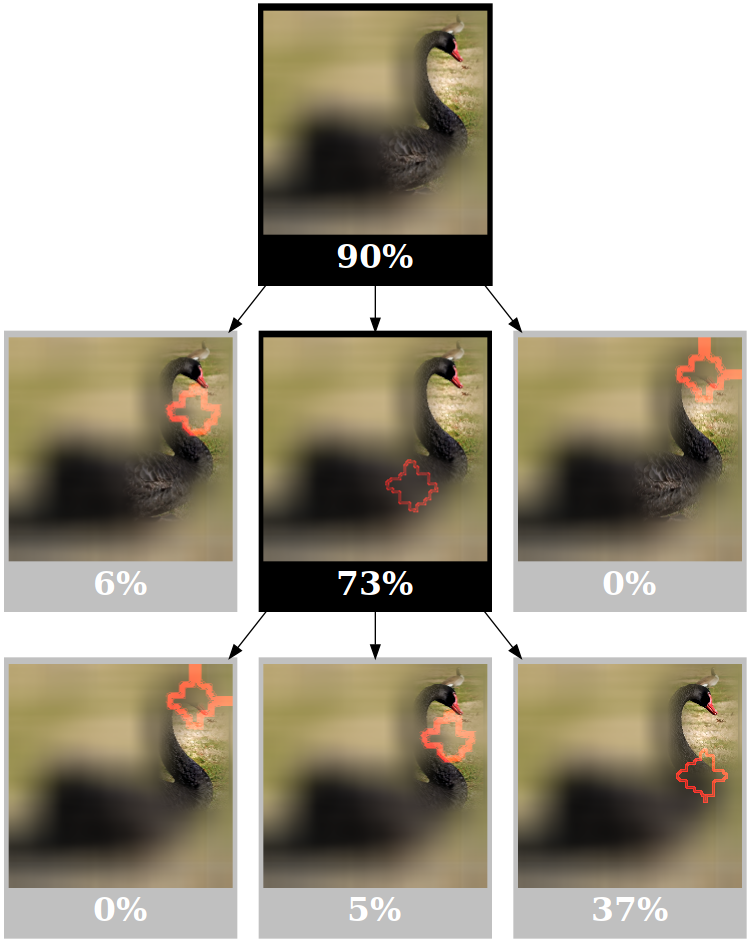}
	\caption{An example SAG tree explaining Fig.~\ref{figure_1044} for VGG.}
	\label{figure_1044_vgg}
\end{figure*}

In Section 4.2, we stated that distilled transformer models sometimes obtain high confidence while only showing a small number of regions. Here we provide more qualitative results from cross-testing, Fig. \ref{figure_igos_example_2}. The \texttt{Cougar} image in the second column of the first row, the heatmap is generated on Swin-T, however, both DeiT-S ($84.68\%$) and DeiT-S-distilled($96.96\%$) have higher confidence than Swin-T ($77.39\%$) on this partially occluded image.

Also, we can see that in the case even if the figure was generated by VGG19, the distilled and non-distilled transformer models have sometimes much higher confidences on occluded images. Especially, DeiT-S-distilled more often has confidence than VGG. This shows the robustness of transformers over convolutional networks.

We also performed 3 trials of cross-testing to generate standard deviations on selected models. Table~\ref{tab:cross_test_seed} shows the quantitative results of cross-testing using the same model with different seeds. We can see that, for the same model, the insertion score of cross-testing results with different seeds has reasonable standard deviations between $0.01$ and with Swin-T (0.052), DeiT-S (0.047) and LeViT-256 (0.041) being the highest. %

\begin{table}[htb]
	\centering
	\resizebox{0.99\linewidth}{!}{%
	\begin{tabular}{l|c|c|c|c|c}
		\toprule
		\bf Model & \bf ResNet50-A2 & \bf Swin-T & \bf DeiT-S & \bf DeiT-S-distilled & \bf LeViT-256 \\
        \midrule
		Del & 0.185 $\pm$ 0.012 & 0.161 $\pm$ 0.021 & 0.168 $\pm$ 0.028 & 0.122 $\pm$ 0.020 & 0.134 $\pm$ 0.028 \\
		Ins & 0.885 $\pm$ 0.010 & 0.869 $\pm$ 0.052 & 0.926 $\pm$ 0.047 & 0.960 $\pm$ 0.019 & 0.931 $\pm$ 0.041 \\
		\bottomrule
	\end{tabular}
	}
 \vskip -0.1in
 \caption{Cross-Testing Results on Models with the same Architecture and Different Seeds. Each column represents the mean and standard deviation of Del/Ins scores obtained by doing cross-testing for models trained using different random seeds with the same architecture.}
\label{tab:cross_test_seed}
\end{table}

\end{document}